\documentclass{article}

\usepackage[accepted]{tmlr}

\usepackage{microtype}
\usepackage{graphicx}
\usepackage{subcaption}
\usepackage{booktabs}
\usepackage{float}

\usepackage{algorithm}
\usepackage{algpseudocode}

\usepackage{tabularx}
\usepackage{array}
\usepackage{multirow}
\usepackage{enumitem}

\usepackage{listings}

\usepackage{xspace}
\usepackage{pifont}
\usepackage[table]{xcolor}

\usepackage[most]{tcolorbox}

\newtcolorbox{prompt}[1][]{
    colback=white,
    colframe=black,
    colbacktitle=black,
    coltitle=white,
    fonttitle=\bfseries\large,
    fontupper=\ttfamily\small,
    arc=1mm,
    breakable,
    #1
}

\newcommand{\CoT}{CoT\xspace}
\newcommand{\ours}{$\pi$-\CoT}
\newcommand{\slice}{\texttt{SLICE}\xspace}
\newcommand{\cmark}{\ding{51}} %
\newcommand{\xmark}{\ding{55}} %

\newcommand{\llama}{Llama-3.3-70B-Instruct\xspace}

\newcommand{\deepseek}{DeepSeek-R1-Distill-Qwen-32B\xspace}

\newcommand{\braces}[1]{\left\{ #1 \right \}}

\def\mbf#1{\mathbf{#1}}

\def\mc#1{\mathcal{#1}}

\def\tbf#1{\textbf{#1}}

\def\balign#1\ealign{\begin{align}#1\end{align}}
\def\baligns#1\ealigns{\begin{align*}#1\end{align*}}
\def\balignat#1\ealign{\begin{alignat}#1\end{alignat}}
\def\balignats#1\ealigns{\begin{alignat*}#1\end{alignat*}}
\def\bitemize#1\eitemize{\begin{itemize}#1\end{itemize}}
\def\benumerate#1\eenumerate{\begin{enumerate}#1\end{enumerate}}

\newenvironment{talign*}
 {\csname align*\endcsname}
 {\endalign}
\newenvironment{talign}
 {\csname align\endcsname}
 {\endalign}

\def\balignst#1\ealignst{\begin{talign*}#1\end{talign*}}
\def\balignt#1\ealignt{\begin{talign}#1\end{talign}}

\definecolor{lightgray}{gray}{0.9}  %
\definecolor{emerald}{HTML}{2ECC71}
\definecolor{amethyst}{HTML}{9B59B6}

\colorlet{emeraldLight}{emerald!15}
\colorlet{amethystLight}{amethyst!15}

\usepackage{hyperref}

\usepackage[utf8]{inputenc}
\usepackage{longtable}
\usepackage{tikz}

\usepackage{amsmath}
\usepackage{amssymb}
\usepackage{mathtools}
\usepackage{amsthm}

\usepackage[capitalize,noabbrev]{cleveref}

\theoremstyle{plain}

\theoremstyle{definition}

\theoremstyle{remark}

\newtheorem{example}{Example}

\usepackage[colorinlistoftodos,textsize=small]{todonotes}

\title{\ours: Prolog-Initialized Chain-of-Thought\\
Prompting for Multi-Hop Question-Answering}

\author{
\name Chao Wan\thanks{Equal contribution.}
\email cw862@cornell.edu \\
\addr Cornell University
\AND
\name Albert Gong\footnotemark[1]
\email agong@cs.cornell.edu \\
\addr Cornell University
\AND
\name Mihir Mishra \email mrm367@cornell.edu \\ 
\addr Cornell University
\AND
\name Carl-Leander Henneking \email ch2273@cornell.edu \\
\addr Cornell University
\AND
\name Claas Beger \email cbb89@cornell.edu \\ 
\addr Cornell University
\AND
\name Kilian Q. Weinberger \email kqw4@cornell.edu \\
\addr Cornell University
}

\def\month{10}
\def\year{2026}
\def\openreview{\url{https://openreview.net/forum?id=gykgXffnE8}}

\begin{document}

\maketitle

\begin{abstract}
Chain-of-Thought (CoT) prompting significantly enhances large language models' (LLMs) problem-solving capabilities, but still struggles with complex multi-hop questions, often falling into circular reasoning patterns or deviating from the logical path entirely.
This limitation is particularly acute in retrieval-augmented generation (RAG) settings, where obtaining the right context is critical.
We introduce \textbf{P}rolog-\textbf{I}nitialized \textbf{C}hain-\textbf{o}f-\textbf{T}hought (\ours), a novel prompting strategy that combines logic programming's structural rigor with language models' flexibility.
\ours reformulates multi-hop questions into Prolog queries decomposed as single-hop sub-queries. These are resolved sequentially, producing intermediate artifacts, with which we initialize the subsequent CoT reasoning procedure.
Extensive experiments demonstrate that \ours matches or outperforms standard RAG and in-context CoT on multi-hop question-answering benchmarks, with the largest gains on questions for which retrieval surfaces the necessary evidence. 
\end{abstract}

\section{Introduction}

Chain-of-thought (\CoT) reasoning has emerged as a powerful paradigm for enhancing the problem-solving capabilities of large language models, substantially improving performance on arithmetic, commonsense, and symbolic reasoning tasks \citep{wei2022chain, kojima2022large}.
By encouraging models to articulate their reasoning process through intermediate steps, \CoT enables more systematic and interpretable problem-solving approaches \citep{zhang2022automatic, wang2022self}.
However, as the complexity of reasoning tasks increases---particularly in multi-hop scenarios where multiple interconnected inferences must be made---CoT systems have been observed to generalize poorly \citep{dziri2023faith} and become trapped in circular reasoning patterns \citep{lo2023hierarchical,yao2023react}.

This limitation becomes especially pronounced in retrieval-augmented generation (RAG) systems, where \CoT excels at single-hop questions that require straightforward document retrieval and reasoning,
but struggles significantly with multi-hop queries that demand the integration of information across multiple sources and reasoning steps \citep{asai2023self}.
The fundamental challenge lies in \CoT's inherent trade-off: while its flexibility allows for creative and adaptive reasoning, this same flexibility can lead to unstructured exploration that fails to maintain logical consistency across complex reasoning chains.

Recent work has explored decomposition-based approaches that break multi-hop questions into manageable single-hop questions \citep{khot2023decomposed, zhou2023least, min2019compositional}. However, even with decomposition, critical gaps remain: models struggle to generate high-quality decompositions without supervision \citep{patel2022question, wolfson2020break}, fail at reliable fact composition across steps \citep{press2023measuring}, and lose track of intermediate state in long reasoning chains \citep{yen2024helmet, liu2024lost}. These failures suggest the need for a more principled reasoning framework that can enforce structure while maintaining flexibility. 

In contrast to natural-language reasoning pioneered by \CoT, the structured reasoning paradigm has been extensively studied for decades in artificial intelligence and logic programming \citep{kowalski1982logic, russell1995modern, mccarthy1960programs, simon1971human}.
Prolog, a declarative programming language explicitly designed for structured reasoning tasks, exemplifies this approach through its systematic query resolution mechanisms and logical rule-based inference \citep{robinson1965machine,kowalski1982logic,clocksin2003programming}. While Prolog's rigid structure ensures logical consistency, it lacks the flexibility to handle ambiguous natural language, cannot easily incorporate unstructured text from documents, and requires precise logical formulations that may not capture the nuanced reasoning needed for real-world questions. 

Recognizing that Prolog and CoT possess complementary strengths and weaknesses,
we introduce \textbf{Prolog-Initialized Chain-of-Thought (\ours)}, a novel prompting strategy that combines the structural rigor of logic programming with the contextual flexibility of natural language reasoning. 
Our approach begins by algorithmically reformulating complex multi-hop reasoning questions into equivalent Prolog queries, where each query is deliberately decomposed into a sequence of single-hop sub-queries.
These sub-queries are then resolved systematically: each is translated into natural language and posed to a RAG \citep{lewis2020retrieval, gao2023retrieval} or in-context CoT system, which retrieves relevant documents and generates answers.
The resulting answers are translated back into Prolog facts and incorporated into the evolving knowledge base.

The key insight underlying $\pi$-CoT is that by structuring the reasoning process through Prolog's query resolution mechanism, we ensure that the retrieved context remains highly relevant. Rather than allowing the model to freely explore the reasoning space, potentially losing track of relevant information or pursuing irrelevant tangents
\citep{yao2023react, dziri2023faith}, 
our approach maintains a structured trajectory that systematically builds toward the final answer. 

At the completion of the Prolog resolution process, we concatenate the original question, all retrieved documents, and the structured Prolog derivation to create a comprehensive context that initializes the final CoT reasoning step. Because most of the heavy-lifting already happens in the initialization process, the final CoT reasoning is far simpler and more successful.

Through extensive experimental evaluation, we demonstrate that \ours is on par or better than traditional RAG and in-context systems on multi-hop question-answering (QA) benchmarks, including HotpotQA, 2WikiMultiHopQA, MuSiQue, and PhantomWiki. Our results suggest that the principled integration of symbolic reasoning structures with neural language models offers a promising direction for developing more reliable and interpretable reasoning systems.

\section{Related Works}
\paragraph{Decomposition for multi-hop question-answering.}
Breaking down a complex problem into smaller, manageable parts is a common technique in LLM prompting \citep{zhou2023least,khot2023decomposed,wei2022chain}.
For open-domain QA, \citet{press2023measuring} prompt the model to generate follow-up questions, and \citet{trivedi2023interleaving} take each new sentence in a CoT as input to the retriever.
Importantly, the language model decomposes the question in natural-language steps.
Recent works have also explored the use of \emph{explicit plans}, usually in the form of Python programs \citep{suris2023vipergpt,khattab2022demonstrate}.
While we do not provide a direct comparison due to different model sizes and/or retrieval setups, Monte Carlo Tree Search \citep{tran2024rare}, test-time scaling \citep{Wang2025ChainofRetrievalAG}, and reinforcement learning methods \citep{li2025search,jin2025search,song2025r1} are emerging as promising approaches to open-domain QA.

\paragraph{Improving language model reasoning with Prolog.}
Many prior works generate Prolog from natural language to improve arithmetic reasoning or multi-hop question-answering.
\citet{wu2025lp,vakharia2024proslm,borazjanizadeh2024reliable} translate questions into Prolog, then query a knowledge base. However, they assume the knowledge base has already been populated with facts. Therefore, they do not address retrieval from unstructured documents.
\citet{tan2024thought,yang2024arithmetic} utilize Prolog as a source of supervised training signals for math and logical reasoning.
\citet{weber2019nlprolog} propose a weak unification strategy in Prolog based on semantic similarity. However, their approach requires training and uses pre-defined predicates extracted from the training text.
The method closest to our work is that of \citet{chen2019neural}. They train an LSTM ``programmer'' to generate programs, which are executed using a BERT-based ``reader'' to produce answers. In this work, we contribute a training-free strategy and demonstrate its effectiveness with recent LLMs like \llama and \deepseek, using both sparse and dense retrievers. The results of \citet{chen2019neural} are also limited by the strictness of a pure Prolog execution.

\paragraph{Fact extraction and summarization.}
Extracting knowledge graph triples from unstructured text is a classical problem in NLP, also known as open information extraction (OpenIE) \citep{angeli2015leveraging,pei2023use,zhou2022survey}. To enhance conventional retrieval techniques, LightRAG \citep{guo2024lightrag} and HippoRAG \citep{jimenez2024hipporag,gutierrez2025rag} demonstrate the effectiveness of fact extraction and GraphRAG \citep{edge2024local} and RAPTOR \citep{sarthi2024raptor} propose methods for clustering and summarization.
Among these methods, HippoRAG 2 from \citet{gutierrez2025rag} performs the best on HotpotQA, 2WikiMultiHopQA, and MuSiQue. We provide a direct comparison of our method to HippoRAG 2 in the results section below.
\section{Preliminaries}
\label{sec:preliminaries}
\newcommand{\query}{\mbf{q}}

Dating back to the 1970s, Prolog\footnote{We specifically use the SWI-Prolog implementation \citep{wielemaker2012swi} of Prolog due to its rich support of aggregation and if-then-else logic, which allows us to resolve questions like ``How many Defense against the Dark Arts teachers have been at Hogwarts?''.} is a powerful way to represent factual knowledge and perform logical inference \citep{colmerauer1996birth,russell1995modern,sterling1994art}.
Solutions in Prolog are verifiable and compositional, making it particularly well-suited for multi-hop question answering where intermediate steps must be chained reliably.

\paragraph{Representing factual knowledge.}
Consider the following English sentence about \emph{Harry Potter} \citep{rowling1997harry}. 
\noindent This statement:
\begin{quote}\small
\textit{Lockhart is a Defense against the Dark Arts teacher at Hogwarts.}
\end{quote}
\noindent can be represented as the following Prolog \textbf{fact}:

\texttt{DADA\_teacher("Lockhart", "Hogwarts").}

In Prolog, \verb+DADA_teacher+ is a \textbf{predicate} and ``Lockhart'' and ``Hogwarts'' are \textbf{assignments}. 
A large amount of real-world knowledge can be encoded in this structured, relational form. In fact, as of early 2025, Wikidata contained over 1.65 billion such knowledge triples.\footnote{https://en.wikipedia.org/wiki/Wikidata}
Throughout the paper, we refer to a collection of Prolog facts as a \textbf{knowledge base}.

\paragraph{Multi-hop questions as Prolog queries.}
Prolog doesn’t just store facts---it also allows us to query them. A \textbf{Prolog query} comprises one or more predicates with unassigned variables and asks whether any satisfying variable assignment exists.
Consider the following example:
\begin{example}
\label{ex:hotpot_example}
\textit{Who is the wife of the Defense Against the Dark Arts teacher at Hogwarts
who is also a werewolf?}
\end{example}

One way of answering this question involves first identifying all the Defense against the Dark Arts teachers at Hogwarts, then filtering for those who are also werewolves, and finally retrieving the wives of the selected people. Each step depends on resolving the previous step(s), and the intermediate space of possible answers can be large.
(Note that J.K. Rowling's books mention seven Defense against the Dark Arts teachers at Hogwarts, so a language model would have to consider seven separate reasoning paths to answer the question.)
A concise way of expressing \cref{ex:hotpot_example} is the following Prolog query:
\begin{equation}
\begin{aligned}
\label{eq:query}
    &\texttt{DADA\_teacher(X, "Hogwarts")}, \\
    &\texttt{werewolf(X)}, \\
    &\texttt{wife(X, Y)},
\end{aligned}
\end{equation}
with \texttt{Y} representing the answer.
Translating questions in natural language to structured queries is a classical problem in the community \citep{zelle1996learning, zettlemoyer2012learning}.

A \textbf{solution} to a Prolog query is a set of variable assignments. 
Assuming we have a pre-populated knowledge base, the solutions are obtained by executing the query.
For example, a knowledge base containing the facts 
\texttt{DADA\_teacher("Lockhart", "Hogwarts")}, 
\texttt{werewolf("Lupin")}, and 
\texttt{wife("Lupin", "Tonks")} 
yields the following solution to query~\cref{eq:query}:
\[
\begin{aligned}
\texttt{X = "Lupin"},\\
\texttt{Y = "Tonks"}.
\end{aligned}
\]
\newcommand{\question}{\text{question}}
\newcommand{\model}{\text{LLM}}
\newcommand{\agent}{A}
\newcommand{\target}{\tau}
\newcommand{\plan}{\text{Plan}}
\newcommand{\literals}{\mc L}
\newcommand{\database}{\text{DB}\xspace}
\newcommand{\state}{S}
\newcommand{\corpus}{\mathcal{C}}

\section{Prolog-Initialized Chain-of-Thought}
\label{sec:method}

\begin{figure}[tb]
    \centering
    \includegraphics[width=0.4\columnwidth]{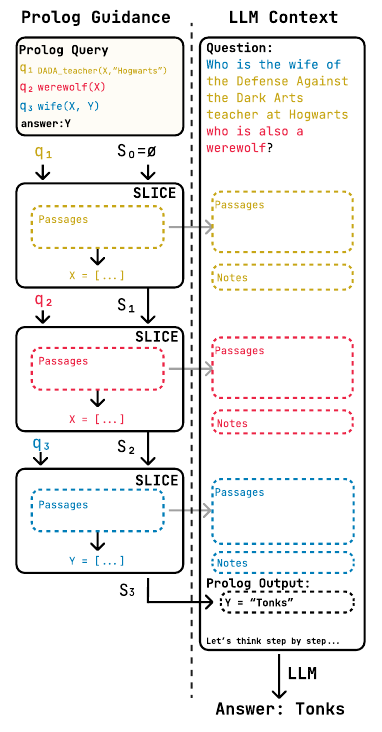}
    \caption{\textbf{Overview of \ours.}
    Left: \ours executes an LLM-generated Prolog query, using the \slice module to resolve each sub-query $q_t$.
    Right: \ours uses the passages, notes, and (potentially) answer from the \slice modules to initialize the CoT prompt for the final LLM call.}
    \label{fig:picot}
\end{figure}

Prolog and LLMs are both powerful tools for reasoning, but they have complementary strengths and weaknesses. 
While Prolog provides a precise and verifiable framework for multi-hop reasoning, it assumes access to a \textit{structured} database, which is often difficult to obtain in practice. In contrast, LLMs can adeptly retrieve and extract information from \textit{unstructured} text, but cannot guarantee logical consistency across reasoning steps.

Motivated by these observations, we combine Prolog and LLMs in two ways: 
a) we initialize the CoT prompt with the execution trace of a Prolog query. 
This mitigates LLMs' tendency to diverge from successful reasoning paths for CoT prompting. 
b) we retrieve relevant information from available documents with the LLM and generate Prolog facts. This provides Prolog an indirect mechanism to interface natural-language documents and create a structured database.

\paragraph{Prolog-Initialized CoT.} 
Our resulting workflow (\ours{}) is illustrated in \cref{fig:picot}. Given the question in natural language, we first prompt\footnote{We provide the query generation prompt in App.~\ref{app:memento_prompts}.} the LLM to generate a structured \textbf{Prolog query}:
\[
\begin{aligned}
\label{eq:prolog_query}
Q = (q_1, q_2, \ldots, q_T).
\end{aligned}
\]
Here, $q_i$ is a \textbf{sub-query}. The query $Q$ can be resolved step-by-step, one sub-query at a time. 
By design, the \textbf{answer} to the original question lies in one of these sub-queries, usually the last one.
For example, the question in \cref{ex:hotpot_example} yields the Prolog query \cref{eq:query} with $T=3$ and answer $\texttt{Y}$.

\ours resolves the query step-by-step and logs the execution trace along the way. 
At step $t$, \ours stores the set of all possible solutions, $\state_t$, to the partially formed Prolog query $(q_1, \ldots, q_t)$. Each element in $\state_t$ is a dictionary of key-value pairs, where the keys are the names of all variables (e.g. \texttt{X}, \texttt{Y}) in the Prolog query and the values are valid assignments (e.g. \texttt{Lupin}).

\subsection{Single-Step Execution with \slice}
\begin{figure}[t]
    \includegraphics[width=\columnwidth]{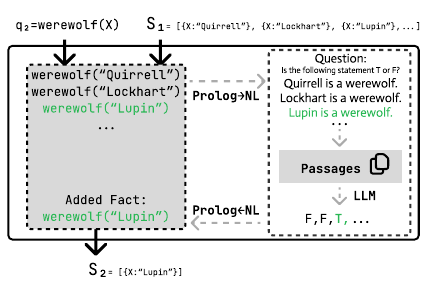}
    \caption{\tbf{\slice module for fact verification in the RAG setting.} At step $t=2$, the module takes in the previous state $S_1$ containing variable assignments, the current sub-query $q_2$, and the corpus $\corpus$ (not shown) as inputs and outputs $\state_2$. Only the Prolog fact (in green) corresponding to a valid statement is added to a growing knowledge base.
    }
    \label{fig:SLICE}
\end{figure}
Each step $t$ follows a fixed procedure, which we refer to as SLICE (\textbf{S}ingle-step \textbf{L}ogical \textbf{I}nference with \textbf{C}ontextual \textbf{E}vidence). 
Given $S_{t-1}$ from the previous iteration, we resolve the next sub-query $q_t$ and create $S_{t}$. 
The sub-query $q_t$ can be of exactly two types: \textbf{Verification} (e.g. \texttt{werewolf(X)}), or \textbf{Extraction} (e.g. \texttt{wife(X,Y)}). Verification queries reduce the set of solutions in $S_{t-1}$ to those compatible with the query (e.g. removing all  teacher names assigned to \texttt{X} that are not werewolves). This scenario is shown in \cref{fig:SLICE} for sub-query $q_2$. 
Extraction queries add new variables to $S_t$ that satisfy the query constraint (e.g. adding variable \texttt{Y} and assigning it the names of all teachers' wives already assigned to \texttt{X}). We resolve both query types with the help of the LLM and the available reference documents, while logging the retrieved document \textit{passages} and Prolog facts (rephrased in natural language as \textit{Notes}). The Prolog facts are stored in a Prolog knowledge base, allowing us to execute the query $(q_1,\dots,q_t)$ with an off-the-shelf Prolog interpreter and log its answer. 
We initialize $S_0=\emptyset$, as the empty set. 
We provide additional details of the \slice module in App.~\ref{app:slice}.

\subsection{\slice Chaining}
The inputs to the \slice module are the sub-query $q_t$, the previous solutions $\state_{t-1}$, and the document corpus $\mc C$. The output of the \slice module is $\state_t$.
To derive the final solution, we chain the \slice modules as follows:
\[
\begin{aligned}
\state_t &= \slice(q_t, \state_{t-1}, \corpus),
&& t = 1,2,\ldots,T, \\
\state_0 &= \emptyset.
\label{eq:slice_chain}
\end{aligned}
\]
By design, the set of solutions is initially empty (i.e., $\state_0 = \emptyset$). As the Prolog execution progresses---and more Prolog facts are collected---$\state_t$ gets closer to the final solution.
After all sub-queries are resolved, $\state_{T}$ is a set of solutions containing the final answer.
For example, \cref{fig:picot} shows $\state_{3} = [\{\texttt{Y}: \texttt{"Tonks"}\}]$ as the final solution.

\subsection{Combining Symbolic and Natural Language Reasoning}
In summary, the iterative process of \slice chaining generates the following artifacts:
\begin{itemize}[leftmargin=10pt, topsep=1pt, itemsep=1pt, parsep=0pt]
    \item a collection of retrieved \textbf{passages} from each \slice execution\footnote{In the in-context setting, the retrieved passages and original passages are the same.};
    \item a collection of Prolog facts, which we convert into natural language (i.e., ``\textbf{notes}''); and
    \item the \textbf{answer} from Prolog execution.
\end{itemize}
The passages contain the necessary factual context to answer the original question.
Since these passages may also contain superfluous information, the notes help to focus the model only on relevant information. Due to their iterative construction, these notes also serve to guide the LLM like breadcrumbs toward the final answer.
When the final solution is non-empty, the LLM merely needs to return it.
For the best results, we provide all three artifacts to the LLM and invoke chain-of-thought reasoning to produce the final answer\footnote{See App.~\ref{app:picot_prompt} for the prompt template.}.

\section{Main Results}

We consider two settings: \tbf{(1) open-domain question-answering}, when the corpus is too large to fit within the model's context window, and \tbf{(2) in-context question-answering}, when the corpus does fit within the model's context window. The open-domain and in-context QA settings are also known as the \emph{fullwiki} and \emph{distractor} settings in the literature \citep{yang2018hotpotqa}. To overcome the context limitations in the open-domain QA setting, we use retrieval augmented generation (RAG) to first fetch relevant passages before generation. Since \ours is a prompting method, we require a strong instruction-tuned model and employ the \llama model from \citet{grattafiori2024llama}. We also provide results utilizing the \deepseek model from \citep{guo2025deepseek} in the in-context question-answering setting.

\subsection{Open-Domain Question-Answering}
\label{sub:fullwiki}
\newcommand{\kdoc}{k}

\begin{table*}[t]
\centering
\caption{\tbf{Accuracy of prompting-based methods on open-domain QA.}
We report mean ± 1 standard error for exact match (EM) and F1 score across 500 randomly chosen questions from each dataset.
Bold indicates that no other method performs significantly better.}
\label{tab:fullwiki}

\begin{tabular}{lcccccc}
\toprule
        & \multicolumn{2}{c}{\textbf{HotpotQA}} & \multicolumn{2}{c}{\textbf{2WikiMultiHopQA}} & \multicolumn{2}{c}{\textbf{MuSiQue}} \\
\cmidrule(lr){2-3} \cmidrule(lr){4-5} \cmidrule(lr){6-7}
\textbf{Method} & EM $\uparrow$ & F1 $\uparrow$ & EM $\uparrow$ & F1 $\uparrow$ & EM $\uparrow$ & F1 $\uparrow$ \\
\midrule
Standard RAG & \textbf{38.8 ± 2.2} & \textbf{52.6 ± 2.0} & 37.2 ± 2.2 & 40.4 ± 2.1 & 11.0 ± 1.4 & 18.1 ± 1.5 \\
Self-Ask & 19.2 ± 1.8 & 28.0 ± 1.8 & 15.8 ± 1.6 & 21.8 ± 1.7 & 5.6 ± 1.0 & 9.1 ± 1.2 \\
IRCoT & \textbf{40.4 ± 2.2} & \textbf{52.9 ± 2.0} & 32.4 ± 2.1 & 42.5 ± 2.0 & \textbf{17.6 ± 1.7} & \textbf{24.5 ± 1.8} \\
\rowcolor{emeraldLight}
\ours (Ours) & \textbf{42.0 ± 2.2} & \textbf{59.1 ± 1.9} & \textbf{49.4 ± 2.2} & \textbf{57.5 ± 2.1} & \textbf{15.2 ± 1.6} & \textbf{25.7 ± 1.7} \\
\bottomrule
\end{tabular}

\vspace{2pt}
{\footnotesize
\emph{Note:} Given a dataset and metric, we perform a repeated measures ANOVA followed by Tukey’s HSD with $\alpha=0.05$ to test significance of paired differences between methods. Omnibus F statistics and the full pairwise comparison matrix are reported in Appendix \ref{app:significance}. 
}

\end{table*}

\begin{table*}
\centering
\caption{\tbf{Efficiency of prompting-based methods on open-domain QA.} We report mean ± 1 standard error number of BM25 queries, LLM calls, and total tokens for the same questions used in \cref{tab:fullwiki}. We further break down the total tokens into prompt and completion tokens in \cref{tab:fullwiki_steps_detailed}.}
\label{tab:fullwiki_steps}

\begin{tabular}{lcccccccccc}
\toprule
& \multicolumn{3}{c}{\textbf{HotpotQA}} & \multicolumn{3}{c}{\textbf{2WikiMultiHopQA}} & \multicolumn{3}{c}{\textbf{MuSiQue}} \\
\cmidrule(lr){2-4} \cmidrule(lr){5-7} \cmidrule(lr){8-10}
\textbf{Method} & BM25 & LLM & Tokens(k) & BM25 & LLM & Tokens(k) & BM25 & LLM & Tokens(k) \\
\midrule
Standard RAG & 1 & 1 & 3.6 & 1 & 1 & 3.7 & 1 & 1 & 2.4 \\
Self-Ask & 3.36 & 3.36 & 15 & 3.44 & 3.44 & 16 & 3.29 & 3.29 & 12 \\
IRCoT & 3.07 & 3.07 & 62 & 3.47 & 3.47 & 59 & 3.51 & 3.51 & 45 \\
\rowcolor{emeraldLight}
\ours (Ours) & 2.82 & 4.82 & 18 & 2.14 & 4.14 & 22 & 3.42 & 5.42 & 15 \\
\bottomrule
\end{tabular}
\end{table*}

We evaluate \ours on three multi-hop QA datasets: (1) HotpotQA from \citet{yang2018hotpotqa}, (2) 2WikiMultiHopQA from \citet{ho2020constructing}, and (3) MuSiQue from \citet{trivedi2022musique}. (PhantomWiki-S/M fit entirely within the context windows of Llama-3.3-70B-Instruct and DeepSeek-R1-Distill-Qwen-32B, so we are not evaluating on them.)
Since these datasets are curated from Wikipedia, we can assess Prolog's effectiveness in handling real-world knowledge.
For our retrieval setup, we use the preprocessed December 18, 2020 corpus from FlashRAG
\citep{jin2025flashrag}, which contains 20M chunks each of size 100 words, and use BM25 \citep{robertson2009probabilistic} as our retriever.
We provide supplementary experiment details in Appendix \ref{app:fullwiki}.

We structure our open-domain QA experiments around two complementary comparison goals.

First, in \cref{tab:fullwiki,tab:fullwiki_steps}, we compare \ours against widely used training-free multi-hop prompting baselines under a shared retrieval setup commonly adopted in prior work. This setting follows the FlashRAG benchmark and uses BM25 retrieval over a large Wikipedia corpus.

Second, in \cref{tab:hipporag}, our goal is to compare \ours against the state-of-the-art GraphRAG-based, OpenIE-augmented retrieval method \textbf{HippoRAG 2}, that outperforms GraphRAG \citep{edge2024local}, RAPTOR \citep{sarthi2024raptor}, and LightRAG \citep{guo2024lightrag} on multi-hop QA. To ensure fidelity to that comparison, we follow the experimental protocol of \citet{gutierrez2025rag}, which uses dense retrieval and curated passage subsets.

Specifically, \cref{tab:fullwiki} compares \ours to standard RAG and two multi-hop RAG baselines that also rely on decomposition (via prompting) to handle multi-hop reasoning: \tbf{(1) Self-Ask} from \citet{asai2023self} and \tbf{(2) IRCoT} from \citep{trivedi2023interleaving}. 
Among all training-free\footnote{Fine-tuned approaches are currently the state-of-the-art on HotpotQA, 2WikiMultiHopQA, and MuSiQue.} methods in the FlashRAG repository, IRCoT was the top-performing training-free method on HotpotQA and the second top-performing method on 2WikiMultiHopQA at the time of writing. 
On HotpotQA, standard RAG, IRCoT, and \ours are comparable in terms of accuracy, with neither method significantly outperforming the other two as determined by a Tukey's HSD test with $\alpha=0.05$. On 2WikiMultiHopQA, \ours significantly outperforms all other methods in terms of exact match and F1 score. On MuSiQue, IRCoT and \ours achieve comparable accuracy.
Despite requiring only a single retriever call, standard RAG achieves surprisingly competitive accuracy on HotpotQA.
\citet{min2019compositional, chen2019understanding} reveal that multi-hop reasoning is not required for many examples in HotpotQA, possibly explaining our findings.
Notably, we find that \ours never performs worse than standard RAG, does significantly better on 2WikiMultiHopQA in both metrics, and significantly better on MuSiQue in F1. Full pairwise tests appear in Appendix \ref{app:significance}.

To assess the efficiency of the methods in \cref{tab:fullwiki}, we measure the number of BM25 queries, the number of LLM calls, and the total token usage. As shown in \cref{tab:fullwiki_steps}, \ours uses a similar number of BM25 queries as IRCoT and Self-Ask.
We also observe \ours using more LLM calls on average than IRCoT and Self-Ask, which makes sense given that \ours must generate a Prolog query and perform the final chain-of-thought reasoning step.
For the price of two extra LLM calls, \ours enjoys lower total token usage. In particular, the use of Prolog allows intermediate steps to use separate, short contexts during execution, rather than a single, long context that grows with the number of steps. Our results in \cref{tab:fullwiki_steps} reflect this fundamental difference. Because wall-clock latency depends on the serving configuration and was measured
on a smaller subset of questions, we report it separately in
Appendix~\ref{app:latency}. These measurements complement the method-level token
and call counts in ~\cref{tab:fullwiki_steps}. 

\begin{table*}[t!]
    \centering
    \caption{\textbf{Comparison to the state-of-the-art OpenIE method.} We report exact match (EM) and F1 on the splits from \citet[Tab.~2]{gutierrez2025rag}. We use \llama for generation and NV-Embed-v2 for embedding passages with $k=5$ per query.}
    \begin{tabular}{l c c c c c c}
        \toprule
        & \multicolumn{2}{c}{\textbf{HotpotQA}} & \multicolumn{2}{c}{\textbf{2WikiMultiHopQA}} & \multicolumn{2}{c}{\textbf{MuSiQue}} \\
        \cmidrule(lr){2-3} \cmidrule(lr){4-5} \cmidrule(lr){6-7}
        \textbf{Method} & EM $\uparrow$ & F1 $\uparrow$ & EM $\uparrow$ & F1 $\uparrow$ & EM $\uparrow$ & F1 $\uparrow$ \\ 
        \midrule
        Standard RAG & 61.2 & 74.5 & 58.3 & 63.2 & 34.9 & 44.8 \\
        HippoRAG 2  & \textbf{62.6} & 75.3 & 65.5 & 72.0 & 37.6 & 49.5 \\
        \rowcolor{emeraldLight}
        \ours (Ours) & 60.3 & \textbf{76.8} & \textbf{71.1}\ddag & \textbf{79.6}\ddag & \textbf{38.5} & \textbf{56.2}\ddag \\
        \bottomrule
    \end{tabular}
    
    \vspace{2mm}
    \begin{minipage}{\textwidth}
    \footnotesize
    \emph{Note:} Due to computational constraints, we report single-run results on 1,000 questions per dataset. Statistical significance is computed using two-proportion $z$-tests: \ddag indicates $p < 0.01$ compared to all baseline methods.
    \end{minipage}
    \label{tab:hipporag}
\end{table*}

Next, we compare \ours to \textbf{HippoRAG 2}, an OpenIE-augmented retrieval method that outperforms GraphRAG \citep{edge2024local}, RAPTOR \citep{sarthi2024raptor}, and LightRAG \citep{guo2024lightrag} on multi-hop QA. We follow the experimental setup of \citet[Tab.~2]{gutierrez2025rag}, which uses the NV-Embed-v2 embedding model of \citet{lee2024nv} for retrieval. Instead of using full Wikipedia as the corpus, this experiment uses 9811 passages for HotpotQA, 6119 passages for 2WikiMultiHopQA, and 11656 passages for MuSiQue.
\cref{tab:hipporag} reports mean exact match and F1 score on the provided 1000 samples for each dataset. These results show \ours outperforming both Standard RAG and HippoRAG 2, demonstrating that offline fact extraction is not necessary for good accuracy on multi-hop question-answering tasks. 

\subsection{In-Context Question-Answering}
\label{sub:incontext}

\begin{figure*}[t]
    \begin{center}
        \includegraphics[width=0.95\textwidth]{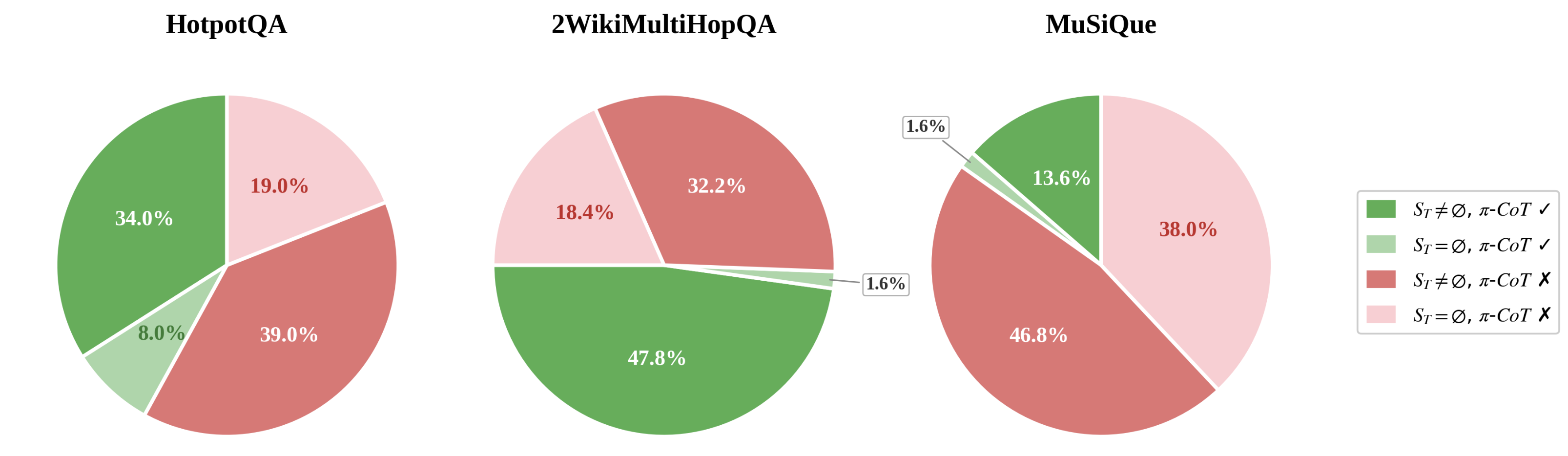}
    \end{center}
\caption{\textbf{Breakdown of \ours predictions by Prolog execution outcome.} Each pie chart shows the distribution of predictions from \cref{tab:fullwiki} across four categories based on whether Prolog returns an answer ($\state_T \neq \emptyset$ vs. $\state_T = \emptyset$) and whether the final answer is correct (green) or incorrect (red). Lighter shades indicate cases where Prolog returns no answer ($\state_T = \emptyset$), while solid colors indicate successful Prolog execution ($\state_T \neq \emptyset$).}
    \label{fig:pi-chart}
\end{figure*}

When all necessary information fits within the model's context window, we investigate when formal reasoning via \ours benefits over natural-language reasoning via CoT. We evaluate on distractor variants of HotpotQA, 2WikiMultiHopQA, and MuSiQue (where gold passages are presented alongside irrelevant distractors), as well as PhantomWiki \citep{gong2025phantomwiki}---a contamination-free benchmark with fictional universes in two variants: PW-S and PW-M (10$\times$ larger corpus). On the Wikipedia-based datasets, \ours performs comparably to CoT with no statistically significant differences (\cref{tab:distractor}), likely because providing gold passages makes the task considerably easier and \llama with CoT approaches a performance ceiling. However, on PhantomWiki, \ours substantially outperforms CoT: on PW-S, F1 increases from 71.9$\pm$1.0\% to 91.4$\pm$0.6\%; on PW-M, from 41.7$\pm$1.1\% to 56.9$\pm$1.0\% (36\% relative gain). Analysis by question difficulty (\cref{fig:ablation}) reveals that \ours maintains accuracy on harder multi-hop questions while CoT degrades, demonstrating \ours's superior ability to track complex multi-branch reasoning chains. Additional details appear in Appendix \ref{app:distractor}. 

\subsection{When does \ours help?}
\label{sec:when}

Table~\ref{tab:coverage} reports \ours's advantage over the baselines as F1 gains in percentage points. Appendix~\ref{app:whentables} provides the full per-dataset breakdowns and an alternative normalization as a robustness check.

\paragraph{Decomposed retrieval reaches evidence a single query misses.}
Restricted to questions where a single BM25 query using the original question
retrieves no gold passage, Table~\ref{tab:recovery} counts how often \ours's
per-sub-query retrieval recovers at least one. Decomposing the query changes what
the retriever is asked for and can surface evidence that retrieval with the original
multi-hop question misses.

\begin{table}[!ht]
\centering
\caption{Questions whose gold evidence a single retrieval query misses, and how often
\ours's decomposed retrieval recovers at least one gold passage.}
\label{tab:recovery}
\begin{tabular}{lrrr}
\toprule
Dataset & Single-shot missed & Recovered & Rate \\
\midrule
HotpotQA & 120 & 37 & 30.8\% \\
2WikiMultiHopQA & 150 & 34 & 22.7\% \\
MuSiQue & 279 & 89 & 31.9\% \\
\bottomrule
\end{tabular}
\end{table}

\paragraph{The advantage tracks evidence access.}
Grouping questions by how much of the annotated gold evidence \ours's retrieval surfaced (Table~\ref{tab:coverage}), we find that the largest gains generally occur when substantial supporting evidence is retrieved, while the gains shrink sharply when little annotated evidence is retrieved. This pattern remains under an alternative normalization reported in Appendix~\ref{app:whentables}. Thus, \ours's advantage is strongly associated with whether its structured pipeline succeeds in surfacing relevant evidence: structure can organize and compose evidence, but it does not substitute for missing evidence.

\begin{table}[!ht]
\centering
\small
\caption{F1 by how much of the annotated gold evidence \ours's retrieval surfaced.
Gains are in percentage points.}
\label{tab:coverage}
\begin{tabular}{llrrrrrr}
\toprule
Dataset & Gold retrieved & $n$ & RAG & IRCoT & \ours & $\Delta$ RAG & $\Delta$ IRCoT \\
\midrule
HotpotQA & all gold passages retrieved & 226 & 64.2 & 62.8 & 73.0 & +8.7 & +10.2 \\
HotpotQA & some retrieved & 180 & 48.9 & 51.1 & 55.7 & +6.8 & +4.6 \\
HotpotQA & none retrieved & 94 & 31.4 & 32.4 & 32.5 & +1.0 & +0.1 \\
2WikiMultiHopQA & all gold passages retrieved & 186 & 59.6 & 63.9 & 79.8 & +20.2 & +15.9 \\
2WikiMultiHopQA & some retrieved & 198 & 36.5 & 38.6 & 56.3 & +19.8 & +17.7 \\
2WikiMultiHopQA & none retrieved & 116 & 16.2 & 15.7 & 23.7 & +7.5 & +8.0 \\
MuSiQue & all gold passages retrieved & 66 & 33.7 & 49.2 & 56.4 & +22.6 & +7.1 \\
MuSiQue & some retrieved & 236 & 19.4 & 23.0 & 28.1 & +8.6 & +5.0 \\
MuSiQue & none retrieved & 198 & 11.2 & 18.0 & 12.7 & +1.5 & -5.3 \\
\bottomrule
\end{tabular}
\end{table}

We additionally examined the available question-structure annotations, but did not
find a consistent relationship between nominal reasoning depth and \ours's
advantage (Appendix~\ref{app:whentables}).

\paragraph{Two caveats.}
First, these strata are defined by \ours's own retrieval, so they characterize
\ours's operating regime rather than an exogenous split; the baselines perform
their own retrieval. We therefore interpret the result as a characterization of when
\ours's advantage is observed, rather than as a controlled comparison. The
recovery analysis above, which is stratified by standard RAG's single-shot retrieval,
does not have this issue.

Second, coverage is measured against each dataset's \emph{annotated} supporting
passages, and these annotations are incomplete. On 2WikiMultiHopQA, among the 25
questions that \ours answers exactly correctly while retrieving none of the
annotated gold titles, 24 have the gold answer string present in passages retrieved
under other titles. This suggests that much of the residual gain in this stratum can
be attributed to evidence retrieved from unannotated articles, rather than to the
complete absence of supporting evidence. The bottom row should therefore be
interpreted as \emph{least annotated evidence} rather than literally \emph{no evidence}.

\section{Ablation Analysis}
\label{sec:ablation}
We analyze how each component in \ours contributes to successfully answering multi-hop questions. Our analysis focuses on two aspects: (1) the role of Prolog-based symbolic reasoning, and (2) the contributions of different information sources in the final chain-of-thought step.

\paragraph{How does Prolog execution contribute to final accuracy?} 
By design, when Prolog execution succeeds in finding an answer, the LLM in the final CoT step is instructed to copy the Prolog solution as its output. This naturally partitions our predictions into four categories based on two binary outcomes: whether Prolog returns a non-empty solution ($\state_T \neq \emptyset$ vs.\ $\state_T = \emptyset$), and whether the final answer is correct (\ours\ \cmark vs.\ \ours\ \xmark). Here, correctness is determined by exact match with the ground-truth answer.

\cref{fig:pi-chart} shows the distribution of predictions from
\cref{tab:fullwiki} across these four cases.
We observe that Prolog successfully returns an answer in the majority of cases:
73.0\% for HotpotQA, 80.0\% for 2WikiMultiHopQA, and 60.4\% for MuSiQue.
Conditional on a non-empty Prolog solution, \ours answers 46.6\%, 59.8\%,
and 22.5\% of these questions correctly, substantially higher than when Prolog
returns no solution (29.6\%, 8.0\%, and 4.0\%, respectively). This strong
association suggests that successful Prolog execution provides a useful signal
for the final prediction.

Interestingly, even when Prolog fails to return an answer ($\state_T = \emptyset$), the artifacts generated during Prolog execution, such as intermediate reasoning notes and partial solutions, can still guide the LLM to the correct answer. This is particularly evident in HotpotQA, where 8.0\% (light green in the leftmost pie in \cref{fig:pi-chart}) of questions are answered correctly despite Prolog returning an empty solution set. This suggests that the structured reasoning process itself, even when incomplete, provides valuable signal for the final answer generation.

For cases where \ours produces incorrect answers (red region, the remaining 27\%, 20\%, and 39.6\% across the three datasets), we provide a detailed failure analysis and categorization in Appendix \ref{app:errors}. The association in \cref{fig:pi-chart} does not by itself isolate the
contribution of the Prolog output. We therefore next ablate the information
provided to the final CoT step.

\paragraph{What information sources are essential for the final CoT step?}
The final chain-of-thought reasoning step (\cref{sec:method}) has access to three sources of information: retrieved passages, reasoning notes from Prolog execution, and the Prolog answer itself. To understand their relative importance, we perform cascading ablations by sequentially removing each component.

\cref{tab:ablation} presents our findings. First, removing passages causes F1 to drop by 4.7\%, 1.2\%, and 2.4\% across the three datasets. This relatively modest drop suggests that most information needed to answer the question is already captured in the reasoning notes and Prolog answer---the retrieved passages primarily serve to fill in gaps or provide additional context.

Next, removing the reasoning notes leads to further drops of 4.8\%, 0.6\%, and 0.8\% F1. The larger impact on HotpotQA (4.8\% vs.\ 0.6--0.8\% on other datasets) indicates that notes are particularly valuable for fuzzy matching when exact string matching in Prolog fails---a common occurrence in HotpotQA which often requires matching entity mentions with slight variations.

Finally, removing the Prolog answer itself causes performance to collapse dramatically on 2WikiMultiHopQA and MuSiQue (to 1.0\% and 4.2\% F1 respectively), while HotpotQA drops to 24.1\% F1. At this point, the model has access only to the question itself. Although we instruct the model to answer based on provided context, the non-zero performance on HotpotQA (24.1\% F1) suggests that \llama still relies on its parametric knowledge for this dataset, likely because HotpotQA questions involve more common knowledge that exists in the model's training data. In contrast, the near-zero performance on 2WikiMultiHopQA (1.0\% F1) and MuSiQue (4.2\% F1) indicates these datasets contain questions that are harder to answer from parametric knowledge alone, making the Prolog-derived reasoning indispensable.

We did not observe challenges with long context when including all passages, notes, and Prolog answers. Therefore, our final model includes all three components to maximize accuracy.

\begin{table}
\centering
\caption{\textbf{Ablation analysis of information sources in the final chain-of-thought step.} We use the experimental setup from \cref{tab:fullwiki}. Components are removed cumulatively from top to bottom.}
\label{tab:ablation} 
\resizebox{\columnwidth}{!}{
\begin{tabular}{lcccccc}
\toprule
        & \multicolumn{2}{c}{\textbf{HotpotQA}} & \multicolumn{2}{c}{\textbf{2WikiMultiHopQA}} & \multicolumn{2}{c}{\textbf{MuSiQue}} \\
        \cmidrule(lr){2-3} \cmidrule(lr){4-5} \cmidrule(lr){6-7}
        \textbf{Ablation} & EM $\uparrow$ & F1 $\uparrow$ & EM $\uparrow$ & F1 $\uparrow$ & EM $\uparrow$ & F1 $\uparrow$ \\
\midrule
\ours (Full) & 42.0 & 59.1 & 49.4 & 57.5 & 15.2 & 25.7 \\[1mm]
\ \ \ w/o Passages & 37.2 & 54.4 & 48.4 & 56.3 & 12.6 & 23.3 \\[1mm]
\ \ \ w/o Notes & 34.0 & 49.6 & 47.8 & 55.7 & 12.4 & 22.5 \\[1mm]
\ \ \ w/o Prolog Answer & 19.2 & 24.1 & 1.0 & 1.0 & 3.2 & 4.2 \\
\bottomrule
\end{tabular}
}
\end{table}
\section{Conclusions \& Future Work}
\label{sec:conclusions}

In this work, we investigate how formal reasoning can guide reasoning in natural language.
We introduce \ours, a novel prompting strategy that initializes the context of an LLM with the intermediate outputs of Prolog-guided execution.
Our results show that even strong LLMs like \llama and \deepseek benefit from being guided through structured reasoning steps, especially on complex, multi-step tasks.
More broadly, our work demonstrates the potential of bringing explicit planning and state tracking into language model behavior.

\paragraph{Robustness to incorrect evidence.}
Our experiments use a fixed Wikipedia corpus and do not explicitly introduce
factually incorrect or systematically contradictory evidence. \ours can
retain multiple extracted answers as separate Prolog facts and reasoning branches
when the evidence supports multiple possibilities. However, this does not itself
resolve genuinely conflicting evidence: if an incorrect or contradictory fact is
accepted by SLICE, it can enter the knowledge base and affect downstream execution.
Source-aware verification and cross-document consistency checking are therefore
important directions for future work.

Despite being a purely prompting-based strategy, \ours has potential implications for training future language models to serve as agents that can piece together knowledge across large, dynamic corpora. Inspired by DeepSeek R1 \citep{guo2025deepseek}, significant effort has been made to couple reasoning with retrieval using reinforcement learning \citep{li2025search,jin2025search,song2025r1}.
An interesting future direction is training language models to generate and reason
through structured queries directly. For example, supervised fine-tuning of smaller
models on \ours-style reasoning traces could help disentangle the benefits of
symbolic scaffolding at inference time from those of learning the structured
reasoning format itself.
\ours provides a way to execute these queries without a pre-existing database.
Some steps in \ours may not even require calling an LLM if the information resides directly in a pre-existing database. Thus, enabling \ours to leverage both unstructured and structured data is a natural next step.

\section*{Statement of Broader Impact}

In this work, we propose a prompting-based method for improving multi-hop question answering by integrating symbolic reasoning structures with large language models. By constraining reasoning through Prolog-initialized, stepwise subqueries, the approach has potential to improve the reliability and consistency of multi-step reasoning. In addition, explicitly decomposing reasoning into structured subqueries and intermediate artifacts may make the reasoning process easier to inspect and retrace, potentially helping users understand how an answer was derived and identify points of failure.

At the same time, as discussed in \cref{sec:ablation}, the use of Prolog introduces its own limitations, including brittleness arising from strict symbolic constraints such as string equality. More broadly, like other advances in language model reasoning, this work could indirectly enable more capable automated systems that may be misused for generating misleading or incorrect information at scale if deployed without appropriate safeguards. The proposed method is purely prompting-based and does not introduce new data sources, training procedures, or deployment mechanisms, and therefore does not inherently alter the risk profile of existing large language models. Responsible use consequently depends on standard practices for evaluation, transparency, and oversight when applying such systems in real-world settings.

\subsubsection*{Acknowledgments}
This work was partially supported by funding from NewYork-Presbyterian for the NYP-Cornell Cardiovascular AI Collaboration and funding from the AI Research Institutes program of the National Science Foundation under Grant Nos. 2433348, 2530143, 2118310, and 2107161. Any opinions, findings, and conclusions or recommendations expressed in this material are those of the author(s) and do not necessarily reflect the views of the National Science Foundation. 
The Authors acknowledge the National Artificial Intelligence Research Resource (NAIRR) Pilot, Purdue Anvil, and SambaNova Cloud for contributing to this research result. We thank Together AI for providing compute credits through the Together AI Research Credits Program. CW thanks Ziyi Yang for assistance with the preparation of Figure 3. 
\bibliography{refs}
\bibliographystyle{tmlr}

\appendix
\onecolumn
\section{Algorithm}

Algorithm~\ref{alg:picot} summarizes the end-to-end execution of \ours.
We provide implementation details of the SLICE module, including parsing and
failure handling, in Appendix~\ref{app:slice}.

\begin{algorithm}[t]
\caption{\ours execution with SLICE chaining}
\label{alg:picot}
\begin{algorithmic}[1]

\Require Question $x$, document corpus $C$
\Ensure Final answer $y$

\State $(Q,\target,D) \gets \textsc{GenerateQuery}(x)$
\State Parse $Q=(q_1,\ldots,q_T)$, target variable $\target$, and definitions $D$
\State Initialize Prolog database $\database \gets \emptyset$
\State Initialize $S_0 \gets \emptyset$, passages $\mathcal{P}\gets\emptyset$,
and notes $\mathcal{N}\gets\emptyset$

\For{$t=1,\ldots,T$}
    \State Instantiate $q_t$ using available assignments in $S_{t-1}$

    \If{$q_t$ contains an unassigned variable}
        \State Convert $q_t$ to an extraction question using $d_t$
    \Else
        \State Convert $q_t$ to a verification question using $d_t$
    \EndIf

    \State Obtain relevant passages $P_t$ from $C$
    \State Query the LLM using the corresponding SLICE prompt

    \If{extraction}
        \State Parse returned values and add corresponding facts to $\database$
    \ElsIf{verification returns true}
        \State Add the instantiated fact to $\database$
    \EndIf

    \State Add $P_t$ and generated facts to $\mathcal{P}$ and $\mathcal{N}$
    \State $S_t \gets \database.\text{query}(q_1,\ldots,q_t)$
\EndFor

\State Extract the Prolog answer $a_{\mathrm{Prolog}}$ from $S_T$
\State Build the final CoT context from $x$, $\mathcal{P}$, $\mathcal{N}$, and $a_{\mathrm{Prolog}}$
\State $y \gets \model(\text{final context})$
\State \Return $y$

\end{algorithmic}
\end{algorithm}

\section{Implementation Details of \slice \ Module}
\label{app:slice}

\paragraph{Inputs.} 
At step $t \in \braces{1,2,\ldots,T}$, \slice\footnote{short for \textbf{S}ingle-step \textbf{L}ogical \textbf{I}nference with \textbf{C}ontextual \textbf{E}vidence} takes as input:
\begin{itemize}[leftmargin=10pt, topsep=1pt, itemsep=1pt, parsep=0pt]
    \item the sub-query, $q_t$;
    \item the solution from the previous step, $\state_{t-1}$; and
    \item the document corpus, $\mathcal{C}$.
\end{itemize}

At step $t=1$, there are no solutions, so $\state_0 = \emptyset$. At step $t>1$, we assign any values in $\state_{t-1}$ to the variables in $q_t$. 
Let's take the second sub-query, $q_2$, from \cref{fig:picot} as an example. 
As shown in \cref{fig:SLICE}, the previous solution $\state_1$ assigns the following values to \texttt{X}: ``Quirrell,'' ``Lockhart,'' ``Lupin,'' etc. These must be substituted into $q_2 = \texttt{werewolf}(\texttt{X})$, yielding the queries \texttt{werewolf("Quirrell")}, \texttt{werewolf("Lockhart")}, \texttt{werewolf("Lupin")}, etc. to resolve for the current step.

While the query \texttt{werewolf("Quirrell")} provides a succinct way of checking whether the claim, ``Quirrell was a werewolf,'' is true, it's still uninterpretable to an LLM. Thus, we need a mechanism to translate Prolog facts to natural-language statements.
Our solution is to prompt\footnote{We provide the definitions generation prompt in \cref{app:memento_prompts}.} the LLM to generate a \textbf{definition} for $q_t$. Each definition comprises two templates:
\begin{itemize}[leftmargin=10pt, topsep=1pt, itemsep=1pt, parsep=0pt]
    \item A \textbf{question template}, which maps $q_t$ to a question (e.g., ``Who are the Defense against the Dark Arts teachers at Hogwarts?'').
    \item A \textbf{statement template}, which maps $q_t$ to a claim (e.g., ``Lockhart is a Defense against the Dark Arts teacher at Hogwarts.'').
\end{itemize}
Each template serves a different purpose, either \emph{fact extraction} or \emph{fact verification}.

\paragraph{Fact extraction.}  
When $q_t$ introduces a new unassigned variable, \slice uses the question template to map $q_t$ to a natural-language question (see \textbf{Prolog$\rightarrow$NL} in \cref{fig:SLICE}). Next, we call the LLM using chain-of-thought prompting to answer this question (see \textbf{LLM} in \cref{fig:SLICE}). 
We provide the LLM prompt in \cref{app:cot_prompt}.
In this work, we consider two strategies to retrieve the relevant evidence for the question:
\begin{enumerate}[leftmargin=20pt,label=(S\arabic*)]
    \item In the \textbf{RAG setting}, we use an external retriever to obtain the top-$k$ passages from $\mathcal{C}$.
    \item In the \textbf{in-context setting}, all passages of $\mathcal{C}$ are provided in the prompt.
\end{enumerate}
Using its innate reading comprehension abilities, the LLM locates the answer to the question from the provided passages and responds with (potentially) multiple answers (e.g., Quirrell, Lockhart, Lupin, etc). These answers are parsed back into Prolog facts (see \textbf{Prolog$\leftarrow$NL} in \cref{fig:SLICE}) and added to the knowledge base.

\paragraph{Fact verification.}  
When all variables in $q_t$ can be assigned values from $\state_{t-1}$, we must check that the fact is true. \slice uses the statement template to generate an entailment question \citep{bowman2015snli}. For example, \texttt{werewolf}(\texttt{"Quirrell"}) corresponds to the question, ``Is the following statement true or false? Quirrell was a werewolf.'' Equipped with S1 or S2, the LLM answers this question (see \textbf{LLM} in \cref{fig:SLICE}).
A true claim is added to the knowledge base as a fact; a false claim is simply ignored (and nothing is added to the knowledge base).
In our running example, the only fact added to the knowledge base at step $t=2$ is \texttt{werewolf}(\texttt{"Lupin"}) (see green text in \cref{fig:SLICE}).

\paragraph{Output.}
Finally, the output of \slice is the updated solution $\state_t$, which can be obtained by querying the knowledge base with $(q_1,\ldots, q_t)$.

\paragraph{Parsing and failure handling.}
For reproducibility, we specify the main implementation details abstracted in
Algorithm~\ref{alg:picot}. Sub-agent outputs are parsed using a restricted literal
grammar supporting quoted strings, numbers, dates, and Boolean values; malformed
extraction outputs fall back to a comma-separated parser, while malformed
verification outputs abort the question. Parsed values are normalized before
assertion into the Prolog knowledge base, and final answers are formatted according
to each dataset's evaluation convention. Generated queries are likewise restricted
to a fixed subset of Prolog syntax; unsupported or malformed queries are rejected
before execution. If an intermediate step returns no solutions, execution continues
where possible without bindings, and the complete conjunction is evaluated after
all steps. Even when Prolog ultimately returns no solution, any retrieved passages
and intermediate notes are still passed to the final CoT step.
Appendix~\ref{app:errors} quantifies the corresponding parsing and execution
failures.

\section{Prompt Templates}

\subsection{Prolog Query and Definitions Generation}
\label{app:memento_prompts}

We use the following prompt template for the Prolog query generation of \cref{sec:method}:

\begin{lstlisting}[basicstyle=\ttfamily,breaklines=true]
You will be provided a question. Your goal is to devise a 
Prolog query to answer this question. Your response must end in 
"**Query:** <query>\n**Target:** <target>\n**Definition:** 
<definition>", where <query> is a Prolog query that when 
executed, will yield the answer to the question, <target> 
is the target variable in the Prolog query to be returned
as the final answer, and <definition> defines the semantic 
meaning of predicates in the Prolog query.

Here are some examples:
(START OF EXAMPLES)
{examples}
(END OF EXAMPLES)

Question: {question}
Answer: 
\end{lstlisting}

To form the prompt, \verb+examples+ is replaced with few-shot examples specific to each dataset and \verb+question+ is replaced with the natural-language question.

\subsection{Chain-of-Thought Fact Extraction and Verification}
\label{app:cot_prompt}

\begin{lstlisting}[basicstyle=\ttfamily,breaklines=true]
You are given the following evidence:
(BEGIN EVIDENCE)
{{evidence}}
(END EVIDENCE)

You will be provided a question. If there is a single answer, 
your response must end with the final answer enclosed in tags: 
<answer>FINAL_ANSWER</answer>
If there are multiple answers, your response must end with the 
final answers enclosed in tags:
<answer>FINAL_ANSWER_1, FINAL_ANSWER_2, ..., FINAL_ANSWER_N</answer>.
If FINAL_ANSWER_N is a string, it must be enclosed in double quotes.
For example, <answer>"FINAL_ANSWER_1", "FINAL_ANSWER_2"</answer>
If FINAL_ANSWER_N is a date, it must be formatted as 
date(year, month, day).
If no information is available to answer the question, 
your response must end with: <answer></answer>.

Here are some examples:
(START OF EXAMPLES)
{{examples}}
(END OF EXAMPLES)

Question: {{question}}
Answer:
\end{lstlisting}

To instantiate the prompt, \verb+evidence+ is replaced by relevant passages, \verb+examples+ is replaced with dataset-specific few-shot examples, and \verb+question+ is replaced with the natural-language question (in the case of fact extraction) or entailment question (in the case of fact verification). Both the instructions and few-shot examples encourage the model to format the answer as Prolog literals so that they can be properly inserted into the Prolog knowledge base.

\subsection{\ours}
\label{app:picot_prompt}

\begin{lstlisting}[basicstyle=\ttfamily,breaklines=true]
You are given the following information:
(BEGIN NOTES)
{{notes}}
(END NOTES)

(BEGIN EVIDENCE)
{{evidence}}
(END EVIDENCE)

You will be provided a question and an answer from a previous 
attempt. If the previous answer is not empty 
(e.g. <answer>...</answer>), you should copy the answer directly. 
If the previous answer is empty (i.e. <answer></answer>), 
you should try to answer the question using 
the notes and evidence provided. If there is a single answer, 
your response must end with the final answer enclosed in tags: 
<answer>FINAL_ANSWER</answer>
If there are multiple answers, your response must end with the 
final answers enclosed in tags:
<answer>FINAL_ANSWER_1,FINAL_ANSWER_2,...,FINAL_ANSWER_N</answer>.
If no information is available to answer the question, 
your response must end with: <answer></answer>.

Here are some examples:
(START OF EXAMPLES)
{{examples}}
(END OF EXAMPLES)
Question: {{question}}
Previous Answer: <answer>{{answer}}</answer>
Answer:
\end{lstlisting}

\section{Supplementary Experiment Details}
\label{app:supplementary}

\subsection{Fullwiki experiment details}
\label{app:fullwiki}

\paragraph{Retrieval setup.} 
We use the \verb+wiki18_100w+ corpus from \url{https://huggingface.co/datasets/RUC-NLPIR/FlashRAG_datasets} and use the code from \url{https://github.com/RUC-NLPIR/FlashRAG} to build our BM25 index.
We allow $k=14$, $k=16$, and $k=8$ chunks per retrieval call for HotpotQA, 2WikiMultiHopQA, and MuSiQue, respectively.

\paragraph{Baseline implementations.}
For standard RAG, we use the Python implementation of \verb+CoTRAGAgent+ from \url{https://github.com/kilian-group/phantom-wiki} and write few-shot examples for each dataset.
For Self-Ask, we use the Python implementation and few-shot examples from \url{https://github.com/RUC-NLPIR/FlashRAG}.
For IRCoT, we use the Python implementation from \url{https://github.com/RUC-NLPIR/FlashRAG} and the GPT3 (\verb+code-davincii-002+) few-shot examples from \url{https://github.com/StonyBrookNLP/ircot} (see also \citep[App.~G]{trivedi2023interleaving}).
We set the maximum iterations for Self-Ask and IRCoT to be 4.

\paragraph{LLM configuration.}
We run \llama on 8 A6000s using vLLM \citep{kwon2023efficient} and use greedy decoding with maximum generation tokens 4096. We use the full 128K context length.

\subsection{Distractor Experiment Details}
\label{app:distractor}

\paragraph{LLM configuration.}
For the \llama results, we use vLLM running on 8 A6000s and use greedy decoding with maximum generation tokens 4096. For the \deepseek results, we use vLLM running on 6 A6000s and use sampling temperature 0.6, top-p 0.95, and max generation tokens 16384. We use the full 128K context length for both models.

\paragraph{PhantomWiki dataset.}
For the experiment of \cref{tab:distractor}(b), we use the code at \url{https://github.com/kilian-group/phantom-wiki} to generate two synthetic multi-hop QA datasets. \cref{tab:phantomwiki-configs} lists the configurations for PW-S and PW-M. Each dataset has 1500 questions.

\begin{table}[h!]
    \centering
    \caption{\tbf{Configurations for PhantomWiki dataset generation.}}
    \begin{tabular}{ccc}
        \toprule
        {\bf Parameter} & {\bf PW-S} & {\bf PW-M}
        \\\midrule
        Question depth & 20 & 20
        \\[1mm]
        Number of family trees & 10 & 100
        \\[1mm]
        Max family tree size & 50 & 50
        \\[1mm]
        Max family tree depth & 20 & 20
        \\[1mm]
        Mode & Easy & Easy
        \\[1mm]
        Number of questions per template & 10 & 10
        \\[1mm]
        Seeds & \{1,2,3\} & \{1,2,3\}
        \\
        \bottomrule
    \end{tabular}
    \label{tab:phantomwiki-configs}
\end{table}

\subsection{LLM Usage Statement}

Large language models were used for proofreading, revising, and literature search. All claims and arguments were drafted and verified by the authors.

\section{Additional Details of Computational Cost}
\label{app:cost}

\subsection{Token and LLM-Call Cost}
\label{app:token-cost}

We provide additional breakdowns of the computational costs reported in the
main text. \Cref{tab:fullwiki_steps_detailed} reports prompt and completion
token usage for open-domain QA, while \cref{tab:distractor-cost} reports token
usage and LLM calls for the in-context setting.

\newcommand{\ptoks}{\textbf{P}\,{\footnotesize\ensuremath{\times 10^3}}}
\newcommand{\ctoks}{\textbf{C}\,{\footnotesize\ensuremath{\times 10^3}}}

\begin{table}[h]
\centering
\caption{\tbf{Token cost of open-domain QA.} We report mean $\pm$ 1 standard error number of prompt tokens \tbf{P} (in thousands) and completion tokens \tbf{C} (in thousands) for the results in \cref{tab:fullwiki_steps}. We use the vLLM inference engine with prefix caching enabled and report the number of cached tokens in parentheses next to the prompt tokens.}
\label{tab:fullwiki_steps_detailed}
\small
\setlength{\tabcolsep}{5pt}
\begin{tabular}{lcccccc}
\toprule
& \multicolumn{2}{c}{\textbf{HotpotQA}} & \multicolumn{2}{c}{\textbf{2WikiMultiHopQA}} & \multicolumn{2}{c}{\textbf{MuSiQue}} \\
\cmidrule(lr){2-3} \cmidrule(lr){4-5} \cmidrule(lr){6-7}
\textbf{Method} & \ptoks & \ctoks & \ptoks & \ctoks & \ptoks & \ctoks \\
\midrule
Standard RAG & 3.46 (0.032) & 0.159 & 3.63 (0.032) & 0.106 & 2.21 (0.073) & 0.136 \\
Self-Ask      & 14.5 (10.5) & 0.402 & 15.7 (10.9) & 0.682 & 11.4 (9.06) & 0.750 \\
IRCoT         & 62.3 (51.5) & 0.047 & 59.2 (44.6) & 0.053 & 45.3 (38.1) & 0.057 \\
\ours         & 18.0 (5.07) & 0.483 & 21.1 (4.21) & 0.543 & 14.4 (3.28) & 0.654 \\
\bottomrule
\end{tabular}
\end{table}

\begin{table}[t]
\centering
\caption{\textbf{Computational cost of in-context QA.}
We report the mean ± 1 standard error number of prompt tokens \tbf{P} (in thousands),
completion tokens \tbf{C} (in thousands), and LLM calls for the results in
\cref{tab:distractor}. We use the vLLM inference engine and report cached tokens
(in parentheses) when prefix caching is enabled.}
\label{tab:distractor-cost}

\small
\setlength{\tabcolsep}{3pt}

\begin{tabular}{lccccccccc}
\toprule
& \multicolumn{3}{c}{\textbf{HotpotQA}} &
  \multicolumn{3}{c}{\textbf{2WikiMultiHopQA}} &
  \multicolumn{3}{c}{\textbf{MuSiQue}} \\
\cmidrule(lr){2-4} \cmidrule(lr){5-7} \cmidrule(lr){8-10}
\textbf{Method} & \ptoks & \ctoks & Calls
& \ptoks & \ctoks & Calls
& \ptoks & \ctoks & Calls \\
\midrule
\multicolumn{10}{l}{\textit{Llama-3.3-70B-Instruct}} \\
CoT   & 2.68 & 0.110 & 1 & 2.10 & 0.0845 & 1 & 3.43 & 0.116 & 1 \\
\ours & 12.4 & 0.412 & 4.37 & 11.0 & 0.437 & 4.47 & 16.1 & 0.553 & 4.90 \\
\midrule[0.3pt]
\multicolumn{10}{l}{\textit{DeepSeek-R1-Distill-Qwen-32B}} \\
CoT   & 2.77 & 0.305 & 1 & 2.20 & 0.298 & 1 & 3.57 & 0.520 & 1 \\
\ours & 11.1 & 1.66 & 4.09 & 9.59 & 1.55 & 4.28 & 12.3 & 1.95 & 4.65 \\
\bottomrule
\end{tabular}

\vspace{0.6em}
\centering\textit{(a) Real-world (Wikipedia-based) multi-hop QA datasets}

\vspace{0.8em}

\begin{tabular}{lccccccc}
\toprule
& \multicolumn{3}{c}{\textbf{PW-S}} &
  \multicolumn{3}{c}{\textbf{PW-M}} \\
\cmidrule(lr){2-4} \cmidrule(lr){5-7}
\textbf{Method} & \ptoks & \ctoks & Calls
& \ptoks & \ctoks & Calls \\
\midrule
\multicolumn{7}{l}{\textit{Llama-3.3-70B-Instruct}} \\
CoT   & 8.12 {\small (8.09)} & 0.400 & 1
      & 68.7 {\small (68.6)} & 0.375 & 1 \\
\ours & 174 {\small (173)} & 1.59 & 23.9
      & 2800 {\small (2754)} & 2.6 & 40 \\
\midrule[0.3pt]
\multicolumn{7}{l}{\textit{DeepSeek-R1-Distill-Qwen-32B}} \\
CoT   & 8.30 {\small (8.26)} & 1.42 & 1
      & 70.8 {\small (70.6)} & 1.13 & 1 \\
\ours & 234 {\small (207)} & 7.28 & 21.0
      & 1260 {\small (1230)} & 7.09 & 16.8 \\
\bottomrule
\end{tabular}

\vspace{0.6em}
\centering\textit{(b) Synthetic multi-hop QA datasets}

\end{table}

\subsection{Wall-Clock Latency}
\label{app:latency}

Token and LLM-call counts characterize computational demand but do not directly
measure end-to-end inference time. We therefore additionally measure wall-clock
latency on the first 50 questions of each open-domain QA split. Because these
measurements are obtained on a smaller subset and under a fixed serving
configuration, we report them separately from the aggregate efficiency results in
Table~\ref{tab:fullwiki_steps}.

All methods process one question at a time without concurrency on
8$\times$A6000 GPUs using Llama-3.3-70B-Instruct. Standard RAG and \ours use
the same vLLM serving configuration, whereas Self-Ask and IRCoT are evaluated
through their FlashRAG implementations with the model loaded in-process. Model
loading is excluded from all measurements. We use batch size one because
wall-clock latency under concurrent serving also depends on scheduling and server
load, whereas our goal here is to measure the end-to-end critical path of a
single question.

\begin{table}[!ht]
\centering
\caption{\textbf{Wall-clock latency of open-domain QA.}
Seconds per question on the first 50 questions of each split, evaluated one
question at a time. We report the mean, median, and maximum because the iterative
baselines exhibit strongly right-skewed runtimes.}
\label{tab:latency}
\small
\setlength{\tabcolsep}{3.5pt}
\begin{tabular}{lrrrrrrrrr}
\toprule
& \multicolumn{3}{c}{\textbf{HotpotQA}}
& \multicolumn{3}{c}{\textbf{2WikiMultiHopQA}}
& \multicolumn{3}{c}{\textbf{MuSiQue}} \\
\cmidrule(lr){2-4} \cmidrule(lr){5-7} \cmidrule(lr){8-10}
\textbf{Method}
& mean & median & max
& mean & median & max
& mean & median & max \\
\midrule
Standard RAG
& 14.8 & 12.6 & 44
& 9.9 & 8.8 & 21
& 10.6 & 8.6 & 28 \\
Self-Ask
& 52.7 & 33.3 & 306
& 448.1 & 56.1 & 5068
& 429.2 & 36.5 & 4854 \\
IRCoT
& 112.9 & 86.6 & 320
& 198.2 & 88.2 & 2993
& 123.0 & 65.5 & 331 \\
\ours (Ours)
& 64.8 & 62.1 & 117
& 65.8 & 62.2 & 94
& 67.2 & 62.5 & 155 \\
\bottomrule
\end{tabular}

\vspace{2pt}
{\footnotesize
All cells are computed over 50 questions except IRCoT on 2WikiMultiHopQA
($n=48$) and MuSiQue ($n=48$). In these cases, two questions raised an error
and produced no answer; failed runs are excluded rather than assigned a latency.}
\end{table}

\paragraph{Latency distribution.}
Standard RAG has the lowest latency across all three datasets. Among the multi-hop methods, median
latencies are of the same order, but their tail behavior differs markedly.
Across all three datasets, \ours has a median latency of approximately
62 seconds, and its slowest question is at most 2.5$\times$ its median. In
contrast, Self-Ask and IRCoT exhibit substantially heavier tails, with the
largest observed ratio between maximum and median latency reaching approximately
133$\times$.

This difference reflects the execution structure of the methods. Once
\ours generates its symbolic plan, it follows the resulting finite sequence
of SLICE operations. Self-Ask and IRCoT instead continue iterative
retrieval--generation until a stopping condition or iteration limit is reached.
Across the three datasets, \ours shows substantially less variation in per-question latency than Self-Ask and IRCoT.

\begin{table}[!ht]
\centering
\caption{\textbf{Latency breakdown of \ours.}
Mean seconds per question by execution stage.}
\label{tab:latency-stages}
\small
\begin{tabular}{lrrr}
\toprule
\textbf{Stage} & \textbf{HotpotQA} & \textbf{2WikiMultiHopQA} & \textbf{MuSiQue} \\
\midrule
Query generation & 23.4 & 24.9 & 22.6 \\
SLICE execution   & 30.6 & 30.5 & 34.1 \\
Final reasoning   & 10.8 & 10.4 & 10.5 \\
\bottomrule
\end{tabular}
\end{table}

\paragraph{Latency breakdown.}
Table~\ref{tab:latency-stages} shows that SLICE execution accounts for the
largest share of \ours's runtime across all three datasets. Verification
operations for distinct candidate assignments are independent and could in
principle be executed concurrently. The measurements reported here use the
sequential implementation evaluated in this work and therefore do not include
any speedup from such parallelization.

\paragraph{Measurement details.}
Wall-clock latency depends on the execution environment, so we report these
results under a fixed hardware and inference setup rather than treating them as
hardware-independent method costs. All measurements use the same model and GPU
configuration, process one question at a time, and exclude model-loading time.
Self-Ask and IRCoT are evaluated through their FlashRAG implementations, while
Standard RAG and \ours use vLLM. The latency measurements therefore
complement the method-level token and LLM-call statistics above by providing an
empirical view of end-to-end inference time under the evaluated setup.
\clearpage
\section{Analysis of \ours Errors}
\label{app:errors}

\subsection{Prolog errors}
\label{app:prolog-errors}

We first analyze cases in which the Prolog execution fails to return a
solution. Table~\ref{tab:error_analysis} breaks these cases down into parsing
and execution errors. These failures account for 27.0\%, 20.0\%, and 39.6\%
of all questions on HotpotQA, 2WikiMultiHopQA, and MuSiQue, respectively.
A complete end-to-end error analysis, including failures outside Prolog
execution, is provided in Appendix~\ref{app:complete-errors}. 

\begin{table}[t]
\centering
\caption{Breakdown of cases in which Prolog fails to return a solution.
Such cases account for 27.0\%, 20.0\%, and 39.6\% of all questions on
HotpotQA, 2WikiMultiHopQA, and MuSiQue, respectively. This is a subset of
the complete \ours error analysis in Appendix~\ref{app:complete-errors}.}
\label{tab:error_analysis}
\small
\setlength{\tabcolsep}{4pt}
\renewcommand{\arraystretch}{1.05}
\begin{tabularx}{\columnwidth}{@{}Xccc@{}}
\toprule
\tbf{Error Type} & \tbf{HotpotQA} & \tbf{2WikiMultiHopQA} & \tbf{MuSiQue} \\
\midrule
\multicolumn{4}{l}{\textit{Parsing errors:}} \\
Prolog query parsing error     & 0.8\% & 0\% & 0.8\% \\
Execution parsing error        & 0.8\% & 0\% & 0.8\% \\
\addlinespace[2pt]
\midrule
\multicolumn{4}{l}{\textit{Execution errors:}} \\
Intermediate predicate existence error & 16.2\% & 19.8\% & 37.4\% \\
Final predicate existence error        & 9.2\% & 0.2\% & 0.6\% \\
\midrule
Total errors & 27.0\% & 20.0\% & 39.6\% \\
\bottomrule
\end{tabularx}
\end{table}

We manually inspected the results in \cref{tab:fullwiki} and identified the following Prolog errors:
\begin{itemize}[leftmargin=10pt]
    \item Prolog Parsing Errors:
    \begin{itemize}[leftmargin=10pt]
        \item \tbf{Prolog query parsing error:} The LLM-generated Prolog query could not be parsed by our Prolog grammar. This error typically occurs due to missing double quotes (see examples in \cref{tab:prolog-parsing-error-1,tab:prolog-parsing-error-2}).
        \item \tbf{Extraction or verification parsing error:} the LLM-generated answer to the fact extraction or fact verification step is an invalid Prolog literal. This error typically occurs due to nested double quotes (see \cref{tab:fact-extraction-parsing-error}).
    \end{itemize}
    \item Prolog Execution Errors:
    \begin{itemize}[leftmargin=10pt]
        \item \tbf{Intermediate predicate existence error:} Information is missing to solve at least one of the intermediate sub-queries.
        \item \tbf{Final predicate existence error:} execution reaches the final step, but no match is returned. For this type, we observe two patterns: 1) genuinely missing information in the final sub-query, 2) mismatches due to Prolog’s strict string equality. The latter can often be resolved when facts are provided back to the model for CoT reasoning. We provide two examples for this type (see examples in \cref{tab:execution-error-1} and \cref{tab:execution-error-2})
    \end{itemize}
\end{itemize}

\begin{table}[t]
\centering
\caption{\tbf{Prolog query parsing error from HotpotQA.} The constant ``...Ready for It?'' is missing double quotes in the Prolog query.}
\label{tab:prolog-parsing-error-1}
\small
\setlength{\tabcolsep}{4pt}
\begin{tabularx}{\columnwidth}{@{}>{\bfseries}l X@{}}
\toprule
\textbf{Question} & ...Ready for It? is a Taylor Swift song from the album \\
                  & scheduled to be released on what date? \\[1mm]
\textbf{Prolog Query} & \texttt{album\_of\_song(...Ready for It?, A1),} \\
                      & \texttt{release\_date(A1, A2)} \\
\bottomrule
\end{tabularx}
\end{table}

\begin{table}
\centering
\caption{\tbf{Prolog query parsing error from MuSiQue.} Negations (\texttt{\textbackslash+}) are not allowed by our Prolog parser.}
\label{tab:prolog-parsing-error-2}
\small
\setlength{\tabcolsep}{4pt}
\begin{tabularx}{\columnwidth}{@{}>{\bfseries}l X@{}}
\toprule
\textbf{Question} & which professional sports team would you not see play a home game \\
                  & in the arena where the last place Cream performed? \\[1mm]
\textbf{Prolog Query} & \texttt{last\_performance\_venue("Cream", A1),} \\
                      & \texttt{all\_professional\_sports\_teams(A3),} \\
                      & \texttt{\textbackslash +home\_teams(A1, A3)} \\
\bottomrule
\end{tabularx}
\end{table}

\begin{table}
\centering
\caption{\tbf{Fact extraction parsing error from HotpotQA.} The double quotes in the generated answer are incorrectly escaped.}
\label{tab:fact-extraction-parsing-error}
\small
\setlength{\tabcolsep}{4pt}
\begin{tabularx}{\columnwidth}{@{}>{\bfseries}l X@{}}
\toprule
\textbf{Question} & What Cantonese slang term can mean both ``ghost man'' \\
                  & and to refer to Westerners? \\[1mm]
\textbf{Response} & ``Gweilo'' or ``gwailou'' \\
\bottomrule
\end{tabularx}
\end{table}

\begin{table}[t]
\centering
\caption{\tbf{Execution error from HotpotQA.}
The generated Prolog query involves the sub-query \texttt{(A1 == A2)} and cannot be satisfied by the facts in the knowledge base.
Although ``Royal Air Force (RAF)'' and ``No.~11 Group RAF'' are semantically related, they are not equal under the Prolog operator \texttt{==}.
The ground-truth answer is \textit{Royal Air Force}.}
\label{tab:execution-error-1}
\small
\setlength{\tabcolsep}{4pt}
\begin{tabularx}{\columnwidth}{@{}>{\bfseries}l X@{}}
\toprule
Question &
What were both Hawker Hurricane and No.~1455 Flight a part of? \\
\midrule
Prolog Query &
\texttt{part\_of("Hawker Hurricane", A1),}\\
&\texttt{part\_of("No. 1455 Flight", A2),}\\
&\texttt{(A1 == A2).} \\
\midrule
Facts &
\texttt{part\_of("Hawker Hurricane", "Royal Air Force (RAF)").}\\
&\texttt{part\_of("Hawker Hurricane", "Royal Yugoslav Air Force (VVKJ)").}\\
&\texttt{part\_of("Hawker Hurricane", "Royal Canadian Air Force").}\\
&\texttt{part\_of("No. 1455 Flight", "No. 11 Group RAF").} \\
\bottomrule
\end{tabularx}
\end{table}

\begin{table}[t]
\centering
\caption{\tbf{Execution error from MuSiQue.} The generated Prolog query involves a string unification \texttt{(A1 = A2)} and cannot be satisfied by the facts in the knowledge base. The ground-truth answer is John D. Loudermilk.}
\label{tab:execution-error-2}
\small
\setlength{\tabcolsep}{4pt}
\begin{tabularx}{\columnwidth}{@{}>{\bfseries}l X@{}}
\toprule
Question &
Who wrote \textit{Turn Me On}, which was performed by the person who also performed \textit{Chasing Pirates}? \\
\midrule
Prolog Query &
\texttt{performer("Chasing Pirates", A1),}\\
&\texttt{performer("turn me on", A2),}\\
&\texttt{A1 = A2 -> writer("turn me on", A3).}\\
\midrule
Facts &
\texttt{performer("Chasing Pirates", "Norah Jones").}\\
&\texttt{performer("turn me on", "Sean Smith").}\\
&\texttt{writer("turn me on", "Greg Lake").}\\
&\texttt{writer("turn me on", "Logan Lynn").}\\
&\texttt{writer("turn me on", "Joni Mitchell").}\\
\bottomrule
\end{tabularx}
\end{table} 

\subsection{Complete error budget}
\label{app:complete-errors}

We next provide a complete error budget over all \ours predictions. We
define an error as a prediction that does not exactly match the reference answer,
consistent with the exact-match (EM) metric used in Table~\ref{tab:fullwiki}. Because
multiple failure conditions can occur in the same example, we assign each incorrect
prediction to the first applicable failure category under a fixed pipeline-order
cascade. The resulting categories are mutually exclusive and therefore sum to the
overall EM error rate.

The Prolog-side categories are aligned with Appendix~\ref{app:prolog-errors}.
Specifically, \emph{Prolog query parsing error} is the same category as in
Appendix~\ref{app:prolog-errors}; \emph{fact extraction / verification parsing
error} corresponds to the second parsing category there; and \emph{predicate
existence error} combines the intermediate and final predicate-existence errors
reported in Table~\ref{tab:error_analysis}. Appendix~\ref{app:prolog-errors}
retains the finer-grained breakdown and qualitative examples.

\begin{table}[!ht]
\centering
\small
\caption{Complete breakdown of \ours predictions counted as incorrect by
exact match. Each prediction is assigned to the first applicable failure category
under a fixed pipeline-order cascade. Percentages are over all 500 questions in each dataset (2Wiki denotes
2WikiMultiHopQA). The final column reports
the range of mean F1 within each category across the three datasets.}
\label{tab:complete-errors}
\begin{tabular}{lrrrr}
\toprule
Failure category & HotpotQA & 2Wiki & MuSiQue & Mean F1 range \\
\midrule
Prolog query parsing error
    & 0.4\% & 0.0\% & 0.2\% & 0 \\
Fact extraction / verification parsing error
    & 0.6\% & 0.0\% & 0.4\% & 0 \\
Predicate existence error
    & 13.2\% & 18.2\% & 35.8\% & 2--14 \\
Agent failure
    & 0.4\% & 0.0\% & 1.2\% & 0--22 \\
Final CoT overrides correct Prolog answer
    & 1.2\% & 3.8\% & 0.2\% & 0--8 \\
No gold passage retrieved
    & 4.8\% & 4.6\% & 16.2\% & 7--13 \\
Gold passage retrieved but required fact not extracted
    & 3.2\% & 0.2\% & 0.4\% & 0--29 \\
Answer correct up to string form
    & 14.2\% & 8.6\% & 9.2\% & 57--65 \\
Substantively incorrect answer
    & 20.0\% & 15.2\% & 21.2\% & 9--27 \\
\midrule
\textbf{Total EM errors}
    & \textbf{58.0\%} & \textbf{50.6\%} & \textbf{84.8\%} & \\
\bottomrule
\end{tabular}
\end{table}

A first observation is that malformed Prolog itself is rare. Prolog query
generation and fact extraction / verification parsing errors together account for
only 1.0\%, 0.0\%, and 0.6\% of all questions on HotpotQA,
2WikiMultiHopQA, and MuSiQue, respectively. The dominant Prolog-side failure is
instead the predicate existence error, particularly on MuSiQue. This suggests that
the principal bottleneck is not generating syntactically executable Prolog, but
populating the resulting predicates with the facts required for execution. We
analyze the relationship between these existence errors and retrieval in
Appendix~\ref{app:retrieval-errors}.

A second substantial source of measured error is answer-form mismatch. Between
8.6\% and 14.2\% of all questions fall into the \emph{answer correct up to string
form} category, with mean F1 between 57 and 65. These predictions contain the
intended answer but fail exact match because of differences in surface form. For
example, a reference answer may be ``Alachua'' while \ours returns
``Alachua County.'' Some cases also reflect idiosyncrasies in the reference
annotation. In one HotpotQA example, the gold answer is the full span
``Carolyn Janice Cherry (born September 1, 1942), better known by the pen name
C. J. Cherryh,'' whereas \ours returns ``C. J. Cherryh.'' These cases suggest
that answer normalization or relaxed string matching could mitigate a nontrivial
fraction of apparent exact-match failures.

\paragraph{Overlapping failure conditions.}
The first-cause analysis above is intentionally mutually exclusive, but failures can
co-occur within the same example. We therefore additionally measure each failure
condition independently among erroneous predictions. These percentages overlap and
need not sum to 100\%. This analysis reveals failures that can be masked by an
earlier category in the cascade. For example, on 2WikiMultiHopQA,
\emph{gold passage retrieved but required fact not extracted} accounts for only
0.4\% of erroneous predictions when assigned as the first cause, but occurs in
36.4\% of erroneous predictions when measured independently. This gap indicates
that fact-population failures often coexist with earlier failures and motivates the
retrieval-conditioned analysis in Appendix~\ref{app:retrieval-errors}. 

\subsection{Errors conditioned on retrieval}
\label{app:retrieval-errors}

The dominant Prolog-side failure in Appendix~\ref{app:complete-errors} is the
predicate existence error, particularly on MuSiQue. We therefore examine how often
this failure occurs depending on whether \ours retrieves the annotated gold
passages required for the question.

\begin{table}[!ht]
\centering
\small
\caption{Probability of a predicate existence error conditioned on whether the
annotated gold passages are retrieved by \ours. The ratio compares the error
rate when gold passages are missed to the error rate when they are retrieved.}
\label{tab:existence-retrieval}
\begin{tabular}{lrrr}
\toprule
 & HotpotQA & 2WikiMultiHopQA & MuSiQue \\
\midrule
Gold passages missed
    & 0.56 & 0.70 & 0.53 \\
Gold passages retrieved
    & 0.12 & 0.17 & 0.34 \\
Ratio
    & 4.6$\times$ & 4.2$\times$ & 1.6$\times$ \\
\bottomrule
\end{tabular}
\end{table}

Predicate existence errors are strongly associated with retrieval failure. On
HotpotQA and 2WikiMultiHopQA, the probability of an existence error is more than
four times higher when the annotated gold passages are missed than when they are
retrieved. This pattern is consistent with the mechanics of \ours: when the
relevant document is not retrieved, the required fact cannot be extracted and
asserted into the Prolog knowledge base, leaving the corresponding predicate
unpopulated and causing execution to fail. Because retrieval success is not
randomly assigned, we interpret this as a strong association with a direct
mechanistic explanation rather than as a controlled causal effect.

MuSiQue shows the same qualitative association with retrieval failure, but the gap is substantially smaller. Even when the annotated gold passages are retrieved, the predicate existence error rate remains 0.34, compared with 0.12 on HotpotQA and 0.17 on 2WikiMultiHopQA. This suggests that MuSiQue poses a harder downstream fact-population problem: retrieving the relevant evidence is often not sufficient for \ours to identify and assert the relation required by the generated Prolog query. Thus, MuSiQue's high existence-error rate appears to reflect both retrieval difficulty and the greater difficulty of converting retrieved evidence into the structured facts needed for execution.

\section{In-context Question-Answering results}
\label{incontext}

\definecolor{customorange}{HTML}{ef8635}
\definecolor{custompurple}{HTML}{8d68b8}

\begin{table*}[h]
\centering
\caption{\textbf{Accuracy of prompting-based methods on in-context QA.} We report exact match (EM) and F1 score on HotpotQA, 2WikiMultiHopQA, MuSiQue, PhantomWiki with a corpus size of 50 articles (PW-S), and PhantomWiki with a corpus size of 500 articles (PW-M). Bold indicates that \ours significantly outperforms CoT ($p<0.05$) according to a paired samples $t$-test given a dataset and metric. We report the computational cost in \cref{tab:distractor-cost}.}

\begin{tabular}{lcccccc}
\toprule
        & \multicolumn{2}{c}{\textbf{HotpotQA}} & \multicolumn{2}{c}{\textbf{2WikiMultiHopQA}} & \multicolumn{2}{c}{\textbf{MuSiQue}} \\
        \cmidrule(lr){2-3} \cmidrule(lr){4-5} \cmidrule(lr){6-7}
        \textbf{Method} & EM $\uparrow$ & F1 $\uparrow$ & EM $\uparrow$ & F1 $\uparrow$ & EM $\uparrow$ & F1 $\uparrow$ \\
\midrule
\rowcolor{custompurple!20} \multicolumn{7}{l}{\textit{Llama-3.3-70B-Instruct}} \\[1mm]
CoT & 62.4 ± 2.2 & 77.4 ± 1.6 & 76.4 ± 1.9 & 83.9 ± 1.5 & 51.2 ± 2.2 & 63.7 ± 1.9 \\[1mm]
\ours & 58.0 ± 2.2 & 77.7 ± 1.5 & 72.8 ± 2.0 & 82.5 ± 1.5 & 46.4 ± 2.2 & 63.1 ± 1.9 \\[1mm]
\midrule[0.3pt]
\rowcolor{customorange!20} \multicolumn{7}{l}{\textit{DeepSeek-R1-Distill-Qwen-32B}} \\[1mm]
CoT & 57.0 ± 2.2 & 74.2 ± 1.7 & 75.2 ± 1.9 & 82.9 ± 1.6 & 45.8 ± 2.2 & 57.3 ± 2.0 \\[1mm]
\ours & 56.8 ± 2.2 & 75.2 ± 1.6 & 74.6 ± 1.9 & 84.2 ± 1.5 & 46.0 ± 2.2 & 59.6 ± 2.0 \\
\bottomrule
\end{tabular}
\\
[2mm]
{(a) Wikipedia-based in-context multi-hop QA datasets}
\\
\vspace{1em}
\begin{tabular}{lcccc}
\toprule
        & \multicolumn{2}{c}{\textbf{PW-S}} & \multicolumn{2}{c}{\textbf{PW-M}} \\
        \cmidrule(lr){2-3} \cmidrule(lr){4-5}
        \textbf{Method} & EM $\uparrow$ & F1 $\uparrow$ & EM $\uparrow$ & F1 $\uparrow$ \\
\midrule
\rowcolor{custompurple!20} \multicolumn{5}{l}{\textit{Llama-3.3-70B-Instruct}} \\[1mm]
CoT & 52.8 ± 1.3 & 71.9 ± 1.0 & 27.5 ± 1.2 & 41.7 ± 1.1 \\[1mm]
\ours & \textbf{78.8 ± 1.1} & \textbf{91.4 ± 0.6} & \textbf{31.1 ± 1.2} & \textbf{56.9 ± 1.0} \\[1mm]
\midrule[0.3pt]
\rowcolor{customorange!20} \multicolumn{5}{l}{\textit{DeepSeek-R1-Distill-Qwen-32B}} \\[1mm]
CoT & 54.4 ± 1.3 & 75.6 ± 0.9 & 17.5 ± 1.0 & 28.0 ± 1.0 \\[1mm]
\ours (Ours) & \textbf{82.7 ± 1.0} & \textbf{88.2 ± 0.8} & 18.4 ± 1.0 & \textbf{31.1 ± 1.0} \\
\bottomrule
\end{tabular}
\\
[2mm]
{(b) Synthetic in-context multi-hop QA datasets}
\label{tab:distractor}
\end{table*}

\begin{figure}
    \begin{center}
        \includegraphics{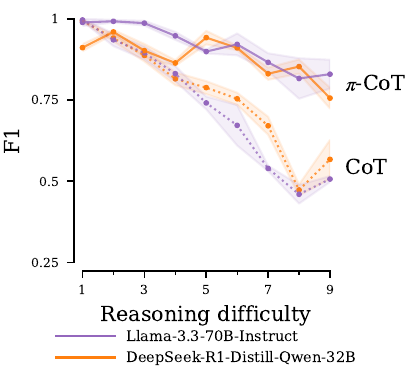}
    \end{center}
    \caption{\tbf{F1 score vs. difficulty, as measured by number of reasoning steps.} We use the synthetic PW-S benchmark from \citet{gong2025phantomwiki} and display mean ± 1 standard error. For each model, we evaluate \CoT and \ours prompting.}
    \label{fig:ablation}
\end{figure}

In this section, we show the results of In-context Question-Answering. 
\cref{tab:distractor}(a) shows that on HotpotQA, 2WikiMultiHopQA, and MuSiQue, 
providing the gold passages to the model makes the task considerably easier than considering all of Wikipedia \citep{min2019compositional}.
Thus, \llama with \CoT may be hitting a performance ceiling.
On PW-S, \llama with \CoT achieves an F1 score of 71.9 ± 1.0\%. This increases to 91.4 ± 0.6\% with \ours. On PW-M, which has a corpus 10 times the size of PW-S, the performance of \llama with \CoT drops to 41.7 ± 1.1\% F1. We posit this drop is mainly due to the inherent challenges of long-context retrieval. On PW-M, \ours significantly boosts the F1 score to 56.9 ± 1.0 F1---a relative gain of 36\%.
In \cref{tab:distractor}(b), we report similar findings using \deepseek as the language model.

Finally, to understand where the gains on PW-S and PW-M come from, we plot accuracy versus the ground-truth difficulty level that comes associated with each question. \citet{gong2025phantomwiki} defines this difficulty level as the number of hops required to answer the question. According to \cref{fig:ablation}, the accuracy of \ours and \CoT is similar for questions with low difficulty. However, \ours diverges from \CoT as the difficulty increases. Since higher difficulty questions require traversing more reasoning paths than lower difficulty questions, our results show that \ours is better able to keep track of multi-hop, multi-branch reasoning than \CoT.

\begin{table}[t]
\centering
\caption{\textbf{Accuracy of CoT with majority voting.}
We report mean exact match (EM) and F1 score on a subset of the datasets from \cref{tab:distractor}.}
\label{tab:distractor-cot-majority}
\small
\setlength{\tabcolsep}{4pt}
\renewcommand{\arraystretch}{1.05}
\begin{tabular}{>{\centering\arraybackslash}p{0.10\columnwidth}cccccccc}
\toprule
{\#}& \multicolumn{2}{c}{\textbf{HotpotQA}} & \multicolumn{2}{c}{\textbf{2WikiMultiHopQA}} &
  \multicolumn{2}{c}{\textbf{MuSiQue}} & \multicolumn{2}{c}{\textbf{PW-S}} \\
\cmidrule(lr){2-3} \cmidrule(lr){4-5} \cmidrule(lr){6-7} \cmidrule(lr){8-9}
\textbf{Samples} & \textbf{EM} $\uparrow$ & \textbf{F1} $\uparrow$ &
\textbf{EM} $\uparrow$ & \textbf{F1} $\uparrow$ &
\textbf{EM} $\uparrow$ & \textbf{F1} $\uparrow$ &
\textbf{EM} $\uparrow$ & \textbf{F1} $\uparrow$ \\
\midrule
\rowcolor{custompurple!20} \multicolumn{9}{l}{\textit{Llama-3.3-70B-Instruct}} \\
1 & 63.0 & 79.8 & 76.8 & 84.2 & 49.6 & 64.3 & 55.73 & 75.43 \\
2 & 63.2 & 80.3 & 77.0 & 84.1 & 48.4 & 62.7 & 54.00 & 73.95 \\
4 & 63.4 & 80.5 & 76.0 & 83.4 & 50.2 & 63.9 & 53.93 & 74.39 \\
8 & 63.2 & 80.2 & 76.0 & 83.3 & 51.0 & 63.8 & 53.13 & 73.83 \\
\midrule[0.3pt]
\rowcolor{customorange!20} \multicolumn{9}{l}{\textit{DeepSeek-R1-Distill-Qwen-32B}} \\
1 & 57.0 & 74.7 & 75.8 & 83.4 & 45.0 & 56.4 & 53.67 & 74.53 \\
2 & 57.0 & 74.8 & 75.6 & 83.2 & 45.6 & 56.2 & 54.73 & 76.14 \\
4 & 57.0 & 75.1 & 76.6 & 83.4 & 46.8 & 57.2 & 54.47 & 75.49 \\
8 & 58.0 & 75.9 & 77.4 & 84.0 & 49.4 & 59.3 & 54.93 & 75.58 \\
\bottomrule
\end{tabular}
\end{table}

We compare standard CoT to the simplest inference-time intervention method: CoT with majority voting. We sample 8 evaluations for each question with the hyperparameters in \cref{tab:sample_hyperparam} and compute the performance when majority voting\footnote{We use the majority definition from \url{https://github.com/EleutherAI/lm-evaluation-harness}.} across 2, 4, and 8 of these samples. On HotpotQA, 2WikiMultiHopQA, and MuSiQue, CoT with majority voting over 4 samples is similar in computational cost to \ours (see \cref{tab:distractor-cost}).
\cref{tab:distractor-cot-majority} shows that the benefits of repeated sampling over single sampling are marginal to none.

\begin{table}[tb]
\centering
\caption{Sampling hyperparameters for the results in \cref{tab:distractor-cot-majority}.}
\label{tab:sample_hyperparam}
\small
\setlength{\tabcolsep}{3pt}
\renewcommand{\arraystretch}{1.05}
\begin{tabularx}{\columnwidth}{@{}
  >{\raggedright\arraybackslash}X
  >{\centering\arraybackslash}p{0.12\columnwidth}
  >{\centering\arraybackslash}p{0.12\columnwidth}
  >{\centering\arraybackslash}p{0.20\columnwidth}
  >{\centering\arraybackslash}p{0.20\columnwidth}
@{}}
\toprule
\textbf{Model} & \textbf{Temp.} & \textbf{Top-p} & \textbf{Rep. Penalty} & \textbf{Max Tokens} \\
\midrule
\llama & 0.6 & 0.9 & 1.0 & 4096 \\
DeepSeek-R1-Distill-Qwen-32B & 0.6 & 0.95 & 1.0 & 16384 \\
\bottomrule
\end{tabularx}
\end{table}

\section{Examples of Generated Prolog Queries and Definitions}
We show five examples of generated Prolog queries and definitions on HotpotQA (\cref{tab:query-hp}), 2WikiMultiHopQA (\cref{tab:query-2wiki}), MuSiQue (\cref{tab:query-msq}), and PhantomWiki (\cref{tab:query-phantomwiki-s}). Examples are randomly chosen. 
\clearpage
\onecolumn
\begin{table}[H]
\centering
\begin{tabular}{p{1.9cm}|p{3.2cm}|c|p{5.8cm}}
\toprule
\tbf{Question} & \tbf{Query} & \multicolumn{1}{r|}{\tbf{Target}} & \tbf{Definitions} \\
\midrule
At what age did Cieli di Toscana's singer become blind? 
&
\parbox[t]{3.2cm}{
singer("Cieli di Toscana", A1)\\
age\_became\_blind(A1, A2)
}
&
A2
&
\parbox[t]{5.8cm}{\scriptsize\ttfamily
singer(<literal1>, <answer>) -> The singer of <literal1> is <answer>. ; Who is the singer of <literal1>?\\[2pt]
age\_became\_blind(<literal1>, <answer>) -> <literal1> became blind at the age of <answer>. ; At what age did <literal1> become blind?
}\\
\midrule
Cooperative Living Organization is located in a city that is the county seat of what county in Florida?
&
\parbox[t]{3.2cm}{
city("Cooperative Living Organization", A1), \\[2pt]
county\_seat(A1, A2)
}
&
A2
&
\parbox[t]{5.8cm}{\scriptsize\ttfamily
- city(<literal1>, <answer>) -> The city where <literal1> is located is <answer>. ; What city is <literal1> located in?\\[2pt]
- county\_seat(<literal1>, <answer>) -> <literal1> is the county seat of <answer>. ; What county is <literal1> the county seat of?
}\\
\midrule
Repertoire of Pluecido Domingo appeared in an opera written by whom which was based on a play by William Shakespeare? 
&
\parbox[t]{3.2cm}{
opera("Repertoire of Pluecido Domingo", A1), \\[2pt]
writer(A1, A2), \\[2pt]
based\_on(A1, A3)
}
&
A2
&
\parbox[t]{5.8cm}{\scriptsize\ttfamily
- opera(<literal1>, <answer>) -> The opera in which <literal1> appeared is <answer>. ; What opera did <literal1> appear in?\\[2pt]
- writer(<literal1>, <answer>) -> The writer of the opera <literal1> is <answer>. ; Who wrote the opera <literal1>?\\[2pt]
- based\_on(<literal1>, <answer>) -> The opera <literal1> was based on a play by William Shakespeare <answer>. ; What play by William Shakespeare was the opera <literal1> based on?
}\\
\midrule
In what 2 countries did both X-Large and A Bathing Ape have stores? 
&
\parbox[t]{3.2cm}{
x\_large\_countries(A1), \\[2pt]
bape\_countries(A2), \\[2pt]
intersection(A1, A2, A3) 
}
&
A3
&
\parbox[t]{5.8cm}{\scriptsize\ttfamily
- x\_large\_countries(<answer>) -> X-Large had stores in <answer> countries. ; In what countries did X-Large have stores?\\[2pt]
- bape\_countries(<answer>) -> A Bathing Ape had stores in <answer> countries. ; In what countries did A Bathing Ape have stores?\\[2pt]
- intersection(<literal1>, <literal2>, <answer>) -> The common countries between <literal1> and <literal2> are <answer>. ; What are the common countries between <literal1> and <literal2>?
}\\
\midrule
Edmund Robert Harris is the principal benefactor of a museum located where ? 
&
\parbox[t]{3.2cm}{
museum("Edmund Robert Harris", A1), \\[2pt]
location(A1, A2)
}
&
A2
&
\parbox[t]{5.8cm}{\scriptsize\ttfamily
- museum(<literal1>, <answer>) -> The museum that <literal1> is the principal benefactor of is <answer>. ; What museum is <literal1> the principal benefactor of?\\[2pt]
- location(<literal1>, <answer>) -> The location of <literal1> is <answer>. ; Where is <literal1> located?
} \\
\bottomrule
\end{tabular}
\caption{5 examples on generated queries and definitions on HotpotQA.}
\label{tab:query-hp}
\end{table}
\clearpage
\onecolumn
\begin{table}[H]
\centering
\begin{tabular}{p{2.0cm}|p{3.2cm}|c|p{5.5cm}}
\toprule
\tbf{Question} & \tbf{Query} & \multicolumn{1}{r|}{\tbf{Target}} & \tbf{Definitions} \\
\midrule
Which film has the director born later, Brutti Di Notte or Bir Türk'E Gönül Verdim?
&
\parbox[t]{3.2cm}{\small
director("Brutti Di Notte", A1),\\
director("Bir Türk'E Gönül Verdim", A2),\\[2pt]
date\_of\_birth(A1, A3),\\[2pt]
date\_of\_birth(A2, A4),\\[2pt]
A3 @> A4 $\rightarrow$ A5
}
&
A5
&
\parbox[t]{5.5cm}{\scriptsize\ttfamily
- director(<literal>, <answer>) -> The director of <literal> is <answer>. ; Who is the director of <literal>?\\[2pt]
- date\_of\_birth(<literal>, <answer>) -> The date of birth of <literal> is <answer>. ; When was <literal> born?
}\\
\midrule
Where did M.~K.~Muthu's father die?
&
\parbox[t]{3.2cm}{\small
father("M. K. Muthu", A1),\\[2pt]
place\_of\_death(A1, A2)
}
&
A2
&
\parbox[t]{5.5cm}{\scriptsize\ttfamily
- father(<literal>, <answer>) -> The father of <literal> is <answer>. ; Who is the father of <literal>?\\[2pt]
- place\_of\_death(<literal>, <answer>) -> The place of death of <literal> is <answer>. ; Where did <literal> die?
}\\
\midrule
What nationality is Julia Parker (Astrologer)'s husband?
&
\parbox[t]{3.2cm}{\small
husband("Julia Parker (Astrologer)", A1),\\[2pt]
nationality(A1, A2)
}
&
A2
&
\parbox[t]{5.5cm}{\scriptsize\ttfamily
- husband(<literal>, <answer>) -> The husband of <literal> is <answer>. ; Who is the husband of <literal>?\\[2pt]
- nationality(<literal>, <answer>) -> The nationality of <literal> is <answer>. ; What is the nationality of <literal>?
}\\
\midrule
Are the directors of both films \emph{The Snake Brothers} and \emph{Kooky} from the same country?
&
\parbox[t]{3.2cm}{\small
director("The Snake Brothers", A1)\\
director("Kooky", A2),\\[2pt]
country\_of\_citizenship(A1, A3),\\[2pt]
country\_of\_citizenship(A2, A4),\\[2pt]
A3 == A4 $\rightarrow$ A5
}
&
A5
&
\parbox[t]{5.5cm}{\scriptsize\ttfamily
- director(<literal>, <answer>) -> The director of <literal> is <answer>. ; Who is the director of <literal>?\\[2pt]
- country\_of\_citizenship(<literal>, <answer>) -> The country of citizenship of <literal> is <answer>. ; What is the country of citizenship of <literal>?
}\\
\midrule
What is the place of birth of the director of film \emph{Kiss The Sky (Film)}?
&
\parbox[t]{3.2cm}{\small
director("Kiss The Sky (Film)", A1),\\[2pt]
place\_of\_birth(A1, A2)
}
&
A2
&
\parbox[t]{5.5cm}{\scriptsize\ttfamily
- director(<literal>, <answer>) -> The director of <literal> is <answer>. ; Who is the director of <literal>?\\[2pt]
- place\_of\_birth(<literal>, <answer>) -> The place of birth of <literal> is <answer>. ; Where was <literal> born?
}\\
\bottomrule
\end{tabular}
\caption{5 examples of generated queries and definitions on 2WikiMultiHopQA.}
\label{tab:query-2wiki}
\end{table}
\clearpage
\onecolumn

\begin{table}[H]
\centering
\begin{tabular}{p{2.0cm}|p{3.2cm}|c|p{5.5cm}}
\toprule
\tbf{Question} & \tbf{Query} & \multicolumn{1}{r|}{\tbf{Target}} & \tbf{Definitions} \\
\midrule
Who was the mother of the person who found the sacred writings that became the Book of Mormon?
&
\parbox[t]{3.2cm}{
founder\_of\_book\_of\\
\_mormon(A1),\\[2pt]
mother(A1, A2)
}
&
A2
&
\parbox[t]{5.5cm}{\scriptsize\ttfamily
- founder\_of\_book\_of\_mormon(<answer>) -> The person who found the sacred writings that became the Book of Mormon is <answer>. ; Who found the sacred writings that became the Book of Mormon?\\[2pt]
- mother(<literal>, <answer>) -> The mother of <literal> is <answer>. ; Who is the mother of <literal>?
}\\
\midrule
Who wrote Turn Me On by the Thinking About You performer?
&
\parbox[t]{3.2cm}{
performer("Thinking About You", A1),\\[2pt]
writer("Turn Me On", A1, A2)
}
&
A2
&
\parbox[t]{5.5cm}{\scriptsize\ttfamily
- performer(<literal>, <answer>) -> The performer of <literal> is <answer>. ; Who performed <literal>?\\[2pt]
- writer(<literal1>, <literal2>, <answer>) -> The writer of <literal1> by <literal2> is <answer>. ; Who wrote <literal1> by <literal2>?
}\\
\midrule
What did M. King Hubbert's employer announce it was in the process of doing in April 2010?
&
\parbox[t]{3.2cm}{
employer("M. King Hubbert", A1),\\[2pt]
announcement(A1, "April 2010", A2)
}
&
A2
&
\parbox[t]{5.5cm}{\scriptsize\ttfamily
- employer(<literal>, <answer>) -> The employer of <literal> is <answer>. ; Who is the employer of <literal>?\\[2pt]
- announcement(<literal>, <date>, <answer>) -> <literal> announced it was in the process of doing <answer> on <date>. ; What did <literal> announce it was doing on <date>?
}\\
\midrule
What is the experimental satellite being forerunner to communication satellite of INSAT-4CR's manufacturer called?
&
\parbox[t]{3.2cm}{
manufacturer("INSAT-\\
4CR", A1),\\[2pt]
experimental\_satellite\\
(A1, A2)
}
&
A2
&
\parbox[t]{5.5cm}{\scriptsize\ttfamily
- manufacturer(<literal>, <answer>) -> The manufacturer of <literal> is <answer>. ; Who is the manufacturer of <literal>?\\[2pt]
- experimental\_satellite(<literal>, <answer>) -> The experimental satellite being the forerunner to the communication satellite of <literal> is <answer>. ; What is the experimental satellite being the forerunner to the communication satellite of <literal>?
}\\
\midrule
The Socialist Autonomous Province of the city where there were mass executions of Danube Swabian population are located in where?
&
\parbox[t]{3.2cm}{
city\_with\_mass\\
\_executions("Danube Swabian", A1),\\[2pt]
socialist\_autonomous\\
\_province(A1, A2),\\[2pt]
location(A2, A3)
}
&
A3
&
\parbox[t]{5.5cm}{\scriptsize\ttfamily
- city\_with\_mass\_executions(<literal>, <answer>) -> The city with mass executions of <literal> population is <answer>. ; What city had mass executions of the <literal> population?\\[2pt]
- socialist\_autonomous\_province(<literal>, <answer>) -> The Socialist Autonomous Province of <literal> is <answer>. ; What is the Socialist Autonomous Province of <literal>?\\[2pt]
- location(<literal>, <answer>) -> <literal> is located in <answer>. ; Where is <literal> located?
}\\
\bottomrule
\end{tabular}
\caption{5 examples on generated queries and definitions on MuSiQue.}
\label{tab:query-msq}
\end{table}
\clearpage
\onecolumn
\begin{table}[H]
\centering
\begin{tabular}{p{2.0cm}|p{3.2cm}|c|p{5.5cm}}
\toprule
\tbf{Question} & \tbf{Query} & \multicolumn{1}{r|}{\tbf{Target}} & \tbf{Definitions} \\
\midrule
How many children does Shelly Reece have?
&
\parbox[t]{3.2cm}{
aggregate\_all(count, distinct(child("Shelly Reece", A1)), A2)
}
&
A2
&
\parbox[t]{5.5cm}{\scriptsize\ttfamily
- child(<literal>, <answer>) -> The child of <literal> is <answer>. ; Who is the child of <literal>?
}\\
\midrule
How many friends does the friend of the person whose hobby is aerospace have?
&
\parbox[t]{3.2cm}{
hobby(A1, "aerospace"),\\[2pt]
friend(A1, A2),\\[2pt]
aggregate\_all(count, distinct(friend(A2, A3)), A4)
}
&
A4
&
\parbox[t]{5.5cm}{\scriptsize\ttfamily
- hobby(<answer>, <literal>) -> The hobby of <answer> is <literal>. ; Who is the person whose hobby is <literal>?\\[2pt]
- friend(<literal>, <answer>) -> The friend of <literal> is <answer>. ; Who is the friend of <literal>?
}\\
\midrule
How many children does the mother of the child of the person whose date of birth is 0985-04-02 have?
&
\parbox[t]{3.2cm}{
dob(A1, "0985-04-02"),\\[2pt]
child(A1, A2),\\[2pt]
mother(A2, A3),\\[2pt]
aggregate\_all(count, distinct(child(A3, A4)), A5)
}
&
A5
&
\parbox[t]{5.5cm}{\scriptsize\ttfamily
- dob(<answer>, <literal>) -> The date of birth of <answer> is <literal>. ; Who is the person whose date of birth is <literal>?\\[2pt]
- child(<literal>, <answer>) -> The child of <literal> is <answer>. ; Who is the child of <literal>?\\[2pt]
- mother(<literal>, <answer>) -> The mother of <literal> is <answer>. ; Who is the mother of <literal>?
}\\
\midrule
How many mothers does the friend of the sister of the sister of the child of the parent of the person whose hobby is mineral collecting have?
&
\parbox[t]{3.2cm}{
hobby(A1, "mineral collecting"),\\[2pt]
parent(A1, A2),\\[2pt]
child(A2, A3),\\[2pt]
sister(A3, A4),\\[2pt]
sister(A4, A5),\\[2pt]
friend(A5, A6),\\[2pt]
aggregate\_all(count, distinct(mother(A6, A7)), A8)
}
&
A8
&
\parbox[t]{5.5cm}{\scriptsize\ttfamily
- hobby(<answer>, <literal>) -> The hobby of <answer> is <literal>. ; Who is the person whose hobby is <literal>?\\[2pt]
- parent(<literal>, <answer>) -> The parent of <literal> is <answer>. ; Who is the parent of <literal>?\\[2pt]
- child(<literal>, <answer>) -> The child of <literal> is <answer>. ; Who is the child of <literal>?\\[2pt]
- sister(<literal>, <answer>) -> The sister of <literal> is <answer>. ; Who is the sister of <literal>?\\[2pt]
- friend(<literal>, <answer>) -> The friend of <literal> is <answer>. ; Who is the friend of <literal>?\\[2pt]
- mother(<literal>, <answer>) -> The mother of <literal> is <answer>. ; Who is the mother of <literal>?
}\\
\midrule
How many children does the sibling of the brother of the brother of the sister of Jamal Song have?
&
\parbox[t]{3.2cm}{
sister("Jamal Song", A1),\\[2pt]
brother(A1, A2),\\[2pt]
brother(A2, A3),\\[2pt]
sibling(A3, A4),\\[2pt]
aggregate\_all(count, distinct(child(A4, A5)), A6)
}
&
A6
&
\parbox[t]{5.5cm}{\scriptsize\ttfamily
- sister(<literal>, <answer>) -> The sister of <literal> is <answer>. ; Who is the sister of <literal>?\\[2pt]
- brother(<literal>, <answer>) -> The brother of <literal> is <answer>. ; Who is the brother of <literal>?\\[2pt]
- sibling(<literal>, <answer>) -> The sibling of <literal> is <answer>. ; Who is the sibling of <literal>?\\[2pt]
- child(<literal>, <answer>) -> The child of <literal> is <answer>. ; Who is the child of <literal>?
}\\
\bottomrule
\end{tabular}
\caption{5 examples on generated queries and definitions on PhantomWiki-S.}
\label{tab:query-phantomwiki-s}
\end{table}
\clearpage
\onecolumn

\section{Examples of execution traces of \ours}
We show the full \ours \ workflow on one example from each of HotpotQA (\autoref{tab:execution-hp}), 2WikiMultiHopQA (\autoref{tab:execution-2wiki}), and MuSiQue (\autoref{tab:execution-msq}). We use the generated predictions from the experiment of \autoref{tab:fullwiki}. Examples are randomly chosen. 
\begin{table}[t]
\centering
\begin{tabular}{p{12.0cm}}
\toprule

\textbf{Question} \\
Did John Updike and Tom Clancy both publish more than 15 bestselling novels? \\

\midrule
\textbf{Query} \\
bestselling\_novels("John Updike", A1), \\
bestselling\_novels("Tom Clancy", A2), \\
(A1 @\textgreater \ 15 → B1 = "Yes" ; B1 = "No"), \\
(A2 @\textgreater \ 15 → B2 = "Yes" ; B2 = "No"), \\
(B1 == "Yes" → (B2 == "Yes" → A3 = "Yes" ; A3 = "No") ; A3 = "No") \\

\midrule
\textbf{Definitions} \\
bestselling\_novels(\textless literal1\textgreater, \textless answer\textgreater) → \textless literal1\textgreater \ published \textless answer\textgreater \ bestselling novels.  
How many bestselling novels did \textless literal1\textgreater \ publish? \\
\midrule

\begin{minipage}[t]{\linewidth}
    \textbf{Sub-Query 1:} How many bestselling novels did John Updike publish?\\[2pt]
    \setlength{\leftskip}{1.5em}%
    \setlength{\parindent}{0pt}%
    \textbf{Retrieved passages:} \textit{John Updike bibliography}, \textit{Tom Wolfe}, \textit{Toward the End of Time},
    \textit{John Updike}, \textit{John Updike} \\
    \textbf{Answer:} more than 20
\end{minipage}

\par\vspace{0.75em}

\begin{minipage}[t]{\linewidth}
    \textbf{Sub-Query 2:} How many bestselling novels did Tom Clancy publish?\\[2pt]
    \setlength{\leftskip}{1.5em}%
    \setlength{\parindent}{0pt}%
    \textbf{Retrieved passages:} \textit{Duty and Honor}, \textit{Line of Sight (novel)}, \textit{Tom Clancy's EndWar},
    \textit{Mike Maden}, \textit{Mark Greaney (novelist)} \\
    \textbf{Answer:} 17
\end{minipage} \\

\midrule

\textbf{Notes} \\
John Updike published more than 20 bestselling novels. \\
Tom Clancy published 17 bestselling novels. \\[6pt]

\textbf{Prolog Answer:} \\
\midrule
\textbf{Final Answer: } Yes \\

\bottomrule
\end{tabular}
\caption{One example on the execution trace of a question on HotpotQA.}
\label{tab:execution-hp}
\end{table}

\begin{table}[t]
\centering
\begin{tabular}{p{12.0cm}}
\toprule
\tbf{Question}  \\
Who is Wisigard's father-in-law?\\
\midrule
\tbf{Query} \\
spouse("Wisigard", A1),\\
father(A1, A2) \\
\midrule
\textbf{Definitions} \\
spouse(\textless literal\textgreater \ , \textless answer\textgreater) → The spouse of \textless literal\textgreater \  is \textless answer\textgreater .  ; Who is the spouse of \textless literal\textgreater ? 
father(\textless literal\textgreater \ , \textless answer\textgreater) → The father of \textless literal\textgreater \  is \textless answer\textgreater . ; Who is the father of \textless literal\textgreater ?\\[2pt]
\midrule

\begin{minipage}[t]{\linewidth}
    \textbf{Sub-Query 1:} Who is the spouse of Wisigard? \\
    \setlength{\leftskip}{1.5em}%
    \setlength{\parindent}{0pt}%
    \textbf{Retrieved passages:} \textit{Wisigard}, \textit{Wisigard}, \textit{Deuteria}, \textit{Deuteria}, \textit{Theudebert I} \\
    \textbf{Answer:} Theudebert I \\
\end{minipage}

\begin{minipage}[t]{\linewidth}
    \textbf{Sub-Query 2:} Who is the father of Theudebert I? \\
    \setlength{\leftskip}{1.5em}%
    \setlength{\parindent}{0pt}%
    \textbf{Retrieved passages:} \textit{Theudebert of Soissons}, \textit{Theudebert of Soissons}, \textit{Theudebert I}, \textit{Theudebert I}, \textit{Theudebert I} \\
    \textbf{Answer:} Theuderic I\\
\end{minipage} \\
\midrule
\textbf{Notes} \\
The spouse of Wisigard is Theudebert I.\\
The father of Theudebert I is Theuderic I.\\[6pt]
\textbf{Prolog Answer:} Theuderic I \\
\midrule
\textbf{Final Answer: } Theuderic I

\\
\bottomrule
\end{tabular}
\caption{One example on the execution trace of a question on 2WikiMultiHopQA.}
\label{tab:execution-2wiki}
\end{table}

\begin{table}[t]
\centering
\begin{tabular}{p{12.0cm}}
\toprule
\tbf{Question}  \\
In what region of the country containing A Luoi is the birthplace of John Phan located? \\
\midrule
\tbf{Query} \\
contains("A Luoi", A1),\\
birthplace("John Phan", A2), \\
region(A1, A2, A3)\\
\midrule
\textbf{Definitions} \\
contains(\textless literal\textgreater, \textless answer\textgreater) → \textless literal\textgreater \ is contained in the country \textless answer\textgreater. ; What country contains \textless literal\textgreater?\\
birthplace(\textless literal\textgreater, \textless answer\textgreater) → The birthplace of \textless literal\textgreater \ is \textless answer\textgreater. ; What is the birthplace of \textless literal\textgreater?\\
region(\textless literal1\textgreater, \textless literal2\textgreater, \textless answer\textgreater) → \textless literal2\textgreater \ is located in the \textless answer\textgreater \ region of \textless literal1\textgreater. ; What region of \textless literal1\textgreater \ is \textless literal2\textgreater \ located in?\\

\midrule

\begin{minipage}[t]{\linewidth}
    \textbf{Sub-Query 1:} What country contains A Luoi? \\
    \setlength{\leftskip}{1.5em}%
    \setlength{\parindent}{0pt}%
    \textbf{Retrieved passages:} \textit{A Luoi Camp}, \textit{A Luoi District}, \textit{A Luoi District}, \textit{A Luoi Camp}, \textit{A Luoi District} \\
    \textbf{Answer:} Vietnam \\
\end{minipage}

\begin{minipage}[t]{\linewidth}
\textbf{Sub-Query 2:}  What is the birthplace of "John Phan"?\\
    \textbf{Retrieved passages:} \textit{John Phan}, \textit{Phan Boi Chau}, \textit{Peter C. Phan}, \textit{John Phan}, \textit{Phan Dinh Phung} \\
    \setlength{\leftskip}{1.5em}%
    \setlength{\parindent}{0pt}%
    \textbf{Answer:} Da Nang, Vietnam \\
\end{minipage}\\

\begin{minipage}[t]{\linewidth}
    \textbf{Sub-Query 3:}  What region of "Vietnam" is "Da Nang, Vietnam" located in? \\
        \setlength{\leftskip}{1.5em}%
    \setlength{\parindent}{0pt}%
    \textbf{Retrieved passages:} \textit{Da Nang}, \textit{Da Nang Air Base}, \textit{Da Nang University of Economics}, \textit{Hoàng Sa District}, \textit{Da Nang} \\
    \textbf{Answer:} South Central Coast \\
\end{minipage}\\

\midrule

\textbf{Notes} \\
A Luoi is contained in the country Vietnam.\\
The birthplace of John Phan is Da Nang, Vietnam.\\
Da Nang, Vietnam is located in the South Central Coast region of Vietnam.\\
\textbf{Prolog Answer: }South Central Coast \\
\midrule
\textbf{Final Answer: }South Central Coast\\

\bottomrule
\end{tabular}
\caption{1 example on the execution trace of a question on MuSiQue.}
\label{tab:execution-msq}
\end{table}

\clearpage
\section{Statistical significance testing}
\label{app:significance}

Table~\ref{tab:fullwiki} reports point estimates with standard errors. Here we
give the significance tests behind them. Because every method answers the same
questions, the comparison is paired: we treat the question as a repeated factor
and run a repeated-measures ANOVA over the four methods
($n=500$ questions per dataset), followed by Tukey's HSD at $\alpha=0.05$ for
all six pairwise comparisons. Tests are run separately per dataset and per
metric. The tests are computed from the same per-question predictions that
produce Table~\ref{tab:fullwiki}; recomputing the means from those predictions
reproduces the published values exactly.

\begin{table}[h]
\centering
\caption{Omnibus repeated-measures ANOVA over the four methods of Table~\ref{tab:fullwiki}. All six tests reject the null hypothesis of equal means.}
\label{tab:anova-omnibus}
\begin{tabular}{llrr}
\toprule
Dataset & Metric & $F(3, 1497)$ & $p$ \\
\midrule
HotpotQA & EM & 52.58 & $2.6\times 10^{-32}$ \\
HotpotQA & F1 & 94.43 & $5.7\times 10^{-56}$ \\
2WikiMultiHopQA & EM & 67.89 & $3.7\times 10^{-41}$ \\
2WikiMultiHopQA & F1 & 83.60 & $5.2\times 10^{-50}$ \\
MuSiQue & EM & 21.13 & $2.1\times 10^{-13}$ \\
MuSiQue & F1 & 41.22 & $1.4\times 10^{-25}$ \\
\bottomrule
\end{tabular}
\end{table}

\begin{table}[h]
\centering
\caption{Tukey HSD pairwise comparisons on HotpotQA, in percentage points, with simultaneous 95\% confidence intervals. Bold indicates significance at $\alpha=0.05$ after correction for multiple comparisons; $p$ values reported as $<$0.001 are upper bounds.}
\label{tab:anova-hotpotqa}
\small
\begin{tabular}{lrcrrcr}
\toprule
& \multicolumn{3}{c}{EM} & \multicolumn{3}{c}{F1} \\
\cmidrule(lr){2-4}\cmidrule(lr){5-7}
Comparison & $\Delta$ & 95\% CI & $p$ & $\Delta$ & 95\% CI & $p$ \\
\midrule
\ours $-$ Standard RAG & +3.2 & [-4.4, +10.8] & 0.679 & +6.6 & [-0.5, +13.6] & 0.078 \\
\ours $-$ IRCoT & +1.6 & [-6.0, +9.2] & 0.900 & +6.2 & [-0.8, +13.3] & 0.104 \\
\ours $-$ Self-Ask & \textbf{+22.8} & [+15.2, +30.4] & $<$0.001 & \textbf{+31.1} & [+24.1, +38.2] & $<$0.001 \\
Standard RAG $-$ IRCoT & -1.6 & [-9.2, +6.0] & 0.900 & -0.3 & [-7.4, +6.7] & 0.900 \\
Standard RAG $-$ Self-Ask & \textbf{+19.6} & [+12.0, +27.2] & $<$0.001 & \textbf{+24.6} & [+17.5, +31.6] & $<$0.001 \\
IRCoT $-$ Self-Ask & \textbf{+21.2} & [+13.6, +28.8] & $<$0.001 & \textbf{+24.9} & [+17.8, +32.0] & $<$0.001 \\
\bottomrule
\end{tabular}
\end{table}

\begin{table}[h]
\centering
\caption{Tukey HSD pairwise comparisons on 2WikiMultiHopQA, in percentage points, with simultaneous 95\% confidence intervals. Bold indicates significance at $\alpha=0.05$ after correction for multiple comparisons; $p$ values reported as $<$0.001 are upper bounds.}
\label{tab:anova-2wiki}
\small
\begin{tabular}{lrcrrcr}
\toprule
& \multicolumn{3}{c}{EM} & \multicolumn{3}{c}{F1} \\
\cmidrule(lr){2-4}\cmidrule(lr){5-7}
Comparison & $\Delta$ & 95\% CI & $p$ & $\Delta$ & 95\% CI & $p$ \\
\midrule
\ours $-$ Standard RAG & \textbf{+12.2} & [+4.8, +19.6] & $<$0.001 & \textbf{+17.1} & [+9.9, +24.3] & $<$0.001 \\
\ours $-$ IRCoT & \textbf{+17.0} & [+9.6, +24.4] & $<$0.001 & \textbf{+14.8} & [+7.6, +22.0] & $<$0.001 \\
\ours $-$ Self-Ask & \textbf{+33.6} & [+26.2, +41.0] & $<$0.001 & \textbf{+35.7} & [+28.5, +42.9] & $<$0.001 \\
Standard RAG $-$ IRCoT & +4.8 & [-2.6, +12.2] & 0.346 & -2.3 & [-9.5, +4.9] & 0.827 \\
Standard RAG $-$ Self-Ask & \textbf{+21.4} & [+14.0, +28.8] & $<$0.001 & \textbf{+18.6} & [+11.4, +25.9] & $<$0.001 \\
IRCoT $-$ Self-Ask & \textbf{+16.6} & [+9.2, +24.0] & $<$0.001 & \textbf{+20.9} & [+13.7, +28.1] & $<$0.001 \\
\bottomrule
\end{tabular}
\end{table}

\begin{table}[h]
\centering
\caption{Tukey HSD pairwise comparisons on MuSiQue, in percentage points, with simultaneous 95\% confidence intervals. Bold indicates significance at $\alpha=0.05$ after correction for multiple comparisons; $p$ values reported as $<$0.001 are upper bounds.}
\label{tab:anova-musique}
\small
\begin{tabular}{lrcrrcr}
\toprule
& \multicolumn{3}{c}{EM} & \multicolumn{3}{c}{F1} \\
\cmidrule(lr){2-4}\cmidrule(lr){5-7}
Comparison & $\Delta$ & 95\% CI & $p$ & $\Delta$ & 95\% CI & $p$ \\
\midrule
\ours $-$ Standard RAG & +4.2 & [-1.1, +9.5] & 0.175 & \textbf{+7.7} & [+2.0, +13.3] & 0.003 \\
\ours $-$ IRCoT & -2.4 & [-7.7, +2.9] & 0.632 & +1.2 & [-4.4, +6.9] & 0.900 \\
\ours $-$ Self-Ask & \textbf{+9.6} & [+4.3, +14.9] & $<$0.001 & \textbf{+16.7} & [+11.0, +22.3] & $<$0.001 \\
Standard RAG $-$ IRCoT & \textbf{-6.6} & [-11.9, -1.3] & 0.008 & \textbf{-6.4} & [-12.1, -0.8] & 0.018 \\
Standard RAG $-$ Self-Ask & \textbf{+5.4} & [+0.1, +10.7] & 0.044 & \textbf{+9.0} & [+3.4, +14.7] & $<$0.001 \\
IRCoT $-$ Self-Ask & \textbf{+12.0} & [+6.7, +17.3] & $<$0.001 & \textbf{+15.5} & [+9.8, +21.1] & $<$0.001 \\
\bottomrule
\end{tabular}
\end{table}

We report intervals alongside $p$ values deliberately. Several comparisons fall
short of the significance threshold while still favouring \ours by a margin
whose interval lies mostly above zero; describing those as evidence of no
difference would overstate what the data support in the opposite direction.

\clearpage
\section{Full when-does-it-help tables}
\label{app:whentables}

This appendix provides the full per-dataset breakdowns supporting Section~\ref{sec:when}, including F1 by retrieval coverage and by question structure. We additionally report an alternative normalization of the coverage gains as a robustness check. 

\begin{table}[!ht]
\centering
\small
\caption{F1 by stratum on HotpotQA, for all four methods. The upper block groups by retrieval coverage, the lower by question structure. Gains are in percentage points.}
\label{tab:strat-hotpotqa}
\begin{tabular}{lrrrrrrr}
\toprule
Stratum & $n$ & RAG & Self-Ask & IRCoT & \ours & $\Delta$ RAG & $\Delta$ IRCoT \\
\midrule
all gold passages retrieved & 226 & 64.2 & 31.7 & 62.8 & 73.0 & +8.7 & +10.2 \\
some retrieved & 180 & 48.9 & 26.9 & 51.1 & 55.7 & +6.8 & +4.6 \\
none retrieved & 94 & 31.4 & 21.0 & 32.4 & 32.5 & +1.0 & +0.1 \\
\midrule
bridge & 406 & 49.1 & 26.0 & 49.3 & 56.3 & +7.2 & +7.0 \\
comparison & 94 & 67.4 & 36.6 & 68.1 & 71.3 & +3.9 & +3.2 \\
\bottomrule
\end{tabular}
\end{table}

\begin{table}[!ht]
\centering
\small
\caption{F1 by stratum on 2WikiMultiHopQA, for all four methods. The upper block groups by retrieval coverage, the lower by question structure. Gains are in percentage points.}
\label{tab:strat-2wiki}
\begin{tabular}{lrrrrrrr}
\toprule
Stratum & $n$ & RAG & Self-Ask & IRCoT & \ours & $\Delta$ RAG & $\Delta$ IRCoT \\
\midrule
all gold passages retrieved & 186 & 59.6 & 31.7 & 63.9 & 79.8 & +20.2 & +15.9 \\
some retrieved & 198 & 36.5 & 17.9 & 38.6 & 56.3 & +19.8 & +17.7 \\
none retrieved & 116 & 16.2 & 12.5 & 15.7 & 23.7 & +7.5 & +8.0 \\
\midrule
compositional & 218 & 30.6 & 21.4 & 44.8 & 46.5 & +16.0 & +1.8 \\
comparison & 119 & 49.7 & 24.0 & 50.6 & 76.9 & +27.1 & +26.3 \\
bridge\_comparison & 112 & 44.8 & 11.8 & 18.0 & 50.2 & +5.3 & +32.1 \\
inference & 51 & 50.9 & 40.4 & 69.6 & 75.1 & +24.3 & +5.5 \\
\bottomrule
\end{tabular}
\end{table}

\begin{table}[!ht]
\centering
\small
\caption{F1 by stratum on MuSiQue, for all four methods. The upper block groups by retrieval coverage, the lower by question structure. Gains are in percentage points.}
\label{tab:strat-musique}
\begin{tabular}{lrrrrrrr}
\toprule
Stratum & $n$ & RAG & Self-Ask & IRCoT & \ours & $\Delta$ RAG & $\Delta$ IRCoT \\
\midrule
all gold passages retrieved & 66 & 33.7 & 15.8 & 49.2 & 56.4 & +22.6 & +7.1 \\
some retrieved & 236 & 19.4 & 10.4 & 23.0 & 28.1 & +8.6 & +5.0 \\
none retrieved & 198 & 11.2 & 5.2 & 18.0 & 12.7 & +1.5 & -5.3 \\
\midrule
2-hop & 282 & 24.1 & 11.4 & 33.2 & 35.3 & +11.2 & +2.1 \\
3-hop & 140 & 9.1 & 4.9 & 15.1 & 14.4 & +5.3 & -0.7 \\
4-hop & 78 & 12.4 & 8.0 & 9.9 & 11.5 & -0.9 & +1.6 \\
\bottomrule
\end{tabular}
\end{table}

\paragraph{Question structure.}
The available question-structure annotations do not reveal a consistent relationship between nominal reasoning depth and \ours's advantage. On MuSiQue, for example, the gain over IRCoT is +2.1, -0.7, and +1.6 F1 for 2-, 3-, and 4-hop questions, respectively. The structural categories in HotpotQA and 2WikiMultiHopQA likewise show no consistent ordering. We therefore do not find evidence that nominal hop count or question structure alone explains when \ours helps.

\paragraph{Alternative normalization.}
Raw percentage-point gains can depend on the baseline's remaining room for improvement. As a robustness check, we additionally normalize each gain by the baseline's remaining headroom, $\Delta/(100-\text{baseline})$. The same broad relationship between evidence coverage and \ours's advantage remains, and the normalized gains decrease monotonically as coverage decreases for all three datasets and both baselines. 

\begin{table}[!ht]
\centering
\small
\caption{Coverage-stratified gains in percentage points and normalized by the baseline's remaining performance headroom. The normalized gains decrease monotonically as retrieval coverage decreases.}
\label{tab:headroom}
\begin{tabular}{llrrrrr}
\toprule
Dataset & Gold retrieved & $n$ & $\Delta$ RAG & Norm. & $\Delta$ IRCoT & Norm. \\
\midrule
HotpotQA & all gold passages retrieved & 226 & +8.7 & +24.4 & +10.2 & +27.3 \\
HotpotQA & some retrieved & 180 & +6.8 & +13.2 & +4.6 & +9.3 \\
HotpotQA & none retrieved & 94 & +1.0 & +1.5 & +0.1 & +0.1 \\
2WikiMultiHopQA & all gold passages retrieved & 186 & +20.2 & +49.9 & +15.9 & +44.0 \\
2WikiMultiHopQA & some retrieved & 198 & +19.8 & +31.2 & +17.7 & +28.9 \\
2WikiMultiHopQA & none retrieved & 116 & +7.5 & +8.9 & +8.0 & +9.5 \\
MuSiQue & all gold passages retrieved & 66 & +22.6 & +34.2 & +7.1 & +14.0 \\
MuSiQue & some retrieved & 236 & +8.6 & +10.7 & +5.0 & +6.5 \\
MuSiQue & none retrieved & 198 & +1.5 & +1.7 & -5.3 & -6.4 \\
\bottomrule
\end{tabular}
\end{table}

\end{document}